\documentclass[journal]{IEEEtran}

 \pdfoutput=1

\usepackage{algpseudocode,algorithm,color}
\newcommand{\transp}{^{\text{\fontfamily{cmss}\fontseries{sbc}\fontshape{n}\selectfont T}}}
\newcommand{\mat}[1]{\boldsymbol{#1}}
\newcommand{\vect}[1]{\boldsymbol{#1}}
\newcommand{\argmin}{\textrm{argmin}}
\newcommand{\argmax}{\textrm{argmax}}

\usepackage{cite}

  \usepackage[pdftex]{graphicx}

\usepackage[cmex10]{amsmath}
\usepackage{amssymb,amsthm}

\newtheorem{remark}{Remark}

\usepackage{array}

\usepackage{stfloats}

\usepackage{url}

\begin{document}
%
\title{A Unified Parallel Algorithm for Regularized Group PLS Scalable to Big Data}
%
%
%

\author{Pierre~{Lafaye de Micheaux}, 
        Beno\^it~Liquet and Matthew Sutton
\thanks{P. Lafaye de Micheaux is with the School of Mathematics and Statistics, University of New South Wales, Sydney, Australia e-mail: lafaye@unsw.edu.au.}
\thanks{B. Liquet is with the Laboratory of Mathematics and its Applications, University of Pau et Pays de L'Adour, UMR CNRS 5142, France and ARC Centre of Excellence for Mathematical and Statistical Frontiers and School of Mathematical
Sciences at Queensland University of Technology, Brisbane, Australia,  e-mail: b.liquet@univ-pau.fr.}
\thanks{M. Sutton is with the ARC Centre of Excellence for Mathematical and Statistical Frontiers and School of Mathematical
Sciences at Queensland University of Technology, Brisbane, Australia, e-mail: m5.sutton@hdr.qut.edu.au.}

}

\maketitle

\begin{abstract}
Partial Least Squares (PLS) methods have been heavily exploited to analyse the association between two blocs of data. These powerful approaches 
can be applied to data sets where the number of variables is greater than the number of observations and in presence of high collinearity between variables.
Different sparse versions of PLS have been developed to integrate multiple data sets while simultaneously selecting the contributing variables. Sparse modelling is a key factor in obtaining better estimators and identifying associations between multiple data sets. The cornerstone of the sparsity version of PLS methods is the link between the SVD of a matrix (constructed from deflated versions of the original matrices of data) and least squares minimisation in linear regression. We present here 
an accurate description of the most popular PLS methods, alongside their mathematical proofs. A unified algorithm is proposed to perform all four types of PLS including their regularised versions.  Various approaches to decrease the computation time are offered, and we show how the whole procedure can be scalable to big data sets.
\end{abstract}


%
\IEEEpeerreviewmaketitle

\section{Introduction}\label{Introduction}

%
%
%
%

 

\IEEEPARstart{I}{n} this article, we review the {\it Partial Least Squares} (PLS) approach to big data. The PLS approach refers to a set of iterative algorithms originally developed by H. Wold \cite{Wold1966}, for the analysis of multiple blocks of data. This article focuses on PLS modelling when there are only two blocks of data. In the two blocks case, the PLS acronym (for Partial Least Squares or Projection to Latent Structures) usually refers to one of four related methods: (i) Partial Least Squares Correlation (PLSC) also called PLS-SVD \cite{Krishnan2011,Abdi2012,Rohlf2000}, (ii) PLS in mode A (PLS-W2A, for Wold's Two-Block, Mode A PLS) 
\cite{Vinzi2010,Cak2016,Wegelin2000}, (iii) PLS in mode B (PLS-W2B) also called Canonical Correlation Analysis (CCA) \cite{Guo2013, Hardoon2004, Hotelling1936}, and (iv) Partial Least Squares Regression (PLS-R, or PLS2) \cite{Wold2001,Rosipal2006,Geladi1986}. The first three methods model a symmetric relationship between the data, aiming to explain the shared correlation or covariance between the datasets, while the fourth method (PLS-R), models an asymmetric relationship, where one block of predictors is used to explain the other block. 

These methods are now widely used in many fields of science, such as genetics \cite{Boulesteix2007,
Ji2011, %
Liquet2016}, neuroimaging \cite{McIntosh1996, %
VanRoon2014} and imaging-genetics \cite{Lorenzi2016, Liu2014}.

  Recently, some authors have started to modify these methods using sparse modelling techniques; see e.g., \cite{LeCao2008a, Dhanjal2009, Witten2009,
Chung2010, Chun2010}. 
These techniques refer to methods in which a relatively small number of covariates have an influence in the model. They are powerful methods  in statistics that provide improved interpretability and better estimators, especially for the analysis of big data. 
For example, in imaging-genetics, sparse models demonstrated great advantages for the identification of biomarkers, leading to more accurate classification of diseases than many existing approaches \cite{Lin2014b}.

In Section~\ref{Section2}, we survey the standard PLS methods. The optimization criteria and algorithmic computation is described. We pay particular attention to the singular value decomposition (SVD) due to its important role in the regularised PLS methods. Gathering in a single document an accurate description of all these methods, alongside with their complete mathematical proofs, constitutes a valuable addition to the literature; see also \cite{Wegelin2000}. A link between the SVD of a matrix (constructed from deflated versions of the original matrices of data) and least squares minimisation in linear regression makes clear how to add regularization to obtain sparsity of the PLS results. This enables us to present in Section~\ref{Section3} the sparse versions of the four types of PLS, as well as a recent group version and a recent sparse group version. An unified algorithm  is then presented in Section~\ref{Section4} to perform all four types of PLS including regularised versions. Various approaches to decrease the computation time are proposed. We explain how the whole procedure can be made scalable to big data sets (any number of measurements, or variables). In Section~\ref{Section5}, we demonstrate the performance of the method on simulated data sets including the case of a categorical response variable. Our algorithm is implemented in the R programming language \cite{R}, and will be maded available on the CRAN as a comprehensive package called \texttt{bigsgPLS} that includes parallel computations.

\section{Partial Least Squares Family }\label{Section2}

\subsection{Notation~\label{Notation}}

Let $\mat{X}: n \times p$ and $\mat{Y}: n \times q$ be two data matrices (or ``blocks'') both consisting of $n$ observations of $p$ and $q$ variables respectively. These variables are generically noted $X$ and $Y$. We assume from now on that these blocks are column-centered (since this turns matrix product into an estimate of covariance, up to a constant factor). Note that scaling is often recommended \cite{Geladi1986}. 
To make explicit the columns of a $n\times r$ matrix $\mat{A}$, we write $\mat{A}:=[\vect{a}_1,\ldots,\vect{a}_r]:=(\vect{a}_j)$. We also note $\mat{A}_{\bullet h}:=[\vect{a}_1,\ldots,\vect{a}_h]$ for the submatrix of the first $h$ columns ($1\leq h\leq r$), and $\mat{A}_{\bullet\bar{h}}:=[\vect{a}_{h+1},\ldots,\vect{a}_r]$ for the remaining ones. 
For two zero-mean vectors $\tilde{\vect{u}}$ and $\tilde{\vect{v}}$ of the same size, we note $Cov(\tilde{\vect{u}}, \tilde{\vect{v}}) = \tilde{\vect{u}}\transp\tilde{\vect{v}}$ and $Cor(\tilde{\vect{u}}, \tilde{\vect{v}}) = \tilde{\vect{u}}\transp\tilde{\vect{v}}/\sqrt{(\tilde{\vect{u}}\transp\tilde{\vect{u}})(\tilde{\vect{v}}\transp\tilde{\vect{v}})}$. (The scaling factor $(n-1)^{-1}$ is omitted w.l.o.g. for a reason that will be made obvious later on, and the tilde symbol is used to emphasize the fact that the vectors are not necessarily normed.) Let $\mat{X}^+$ be the Moore-Penrose (generalized) inverse of $\mat{X}$. We note $\mathcal{P}_{\mat{X}} = \mat{X}\mat{X}^+$ the orthogonal projection matrix onto $\mathcal{I}(\mat{X})$, the space spanned by the columns of $\mat{X}$, and $\mathcal{P}_{\mat{X}^{\perp}} = \mat{I} - \mathcal{P}_{\mat{X}}$ the orthogonal projection matrix on the space orthogonal to $\mathcal{I}(\mat{X})$. When the inverse of $\mat{X}\transp\mat{X}$ exists, we have $\mat{X}^+ = (\mat{X}\transp\mat{X})^{-1}\mat{X}\transp$. The $L_p$ vector norm ($p=1,2$) of an $n$-length vector $\vect{x}$, is $\|\vect{x}\|_p^{} = \left(\sum_{i=1}^n|x_i|^p\right)^{1/p}$. The Frobenius norm of a $n\times r$ matrix $\mat{A}$ is $\|\mat{A}\|_F^{}=\|\texttt{vec}(\mat{A})\|_2$, where the $\texttt{vec}$ operator transforms $\mat{A}$ into an $nr\times1$ vector by stacking its columns. The soft thresholding function is $g^{\textrm{soft}}(x,\lambda)=\textrm{sign}(x)(|x|-\lambda)_{+}$, where $(a)_+=\max(a,0)$. Finally, $\otimes$ denotes the Kronecker product \cite[(3), p.~662]{Lutkepohl2005}.

\subsection{Singular Value Decomposition}
In all four PLS cases, the main linear algebra tool used is the singular value decomposition (SVD). For a real-valued matrix $\mat{M}:p\times q$ of rank $r$, the (full) SVD is given by:
\begin{equation}\label{SVD}
\mat{M} =  \mat{U}\mat{\Delta}\mat{V}\transp = \sum_{l=1}^r\delta_l\vect{u}_l\vect{v}_l\transp,
\end{equation}
where $\mat{U} = (\vect{u}_l):p\times p$ and $\mat{V} = (\vect{v}_l):q\times q$ are two orthogonal matrices whose columns contain the orthonormal left (resp. right) singular vectors, and $\mat{\Delta}=\textrm{diag}(\delta_1,\ldots,\delta_r,0,\ldots,0):p\times q$ is a rectangular matrix containing the corresponding ordered singular values $\delta_1\geq \delta_2\geq\cdots\geq \delta_r>0$. 

Some properties of the SVD will be useful. First, for either orthogonal matrix $\mat{U}$ or $\mat{V}$ and any $h=1,\ldots,r$, we have \[\mat{U}_{\bullet h}\transp\mat{U}_{\bullet h} = \mat{V}_{\bullet h}\transp\mat{V}_{\bullet h} = \mat{I}_h\] 
where $\mat{I}_h$ is the identity matrix. Note that in general $\mat{U}_{\bullet h}\mat{U}_{\bullet h}\transp \neq \mat{I}_p$ unless $h=p$, and $\mat{V}_{\bullet h}\mat{V}_{\bullet h}\transp \neq \mat{I}_q$ unless $h = q$. Second, for $k<r$, the SVD of $\mat{M}-\sum_{l=1}^k\delta_l\vect{u}_l\vect{v}_l\transp$ is $\sum_{l=k+1}^r\delta_l\vect{u}_l\vect{v}_l\transp$. This is called the \textit{SVD deflation property} and it will be used later on in an iterative manner.

Another important property of the SVD states that the (truncated) SVD of $\mat{M}$ provides its best reconstitution (in a least squares sense) by a matrix with a lower rank ($k$, say) \cite[Theorem 21.12.4]{Harville1997}:
$$
\left(\underset{\mat{A}\text{ of rank }k}{\min}\left\|\mat{M} - \mat{A}\right\|_F^2\right)   =  \left\|\mat{M} - \sum_{l=1}^k\delta_l\vect{u}_l\vect{v}_l\transp\right\|_F^2 =  \sum_{l=k+1}^{r}\delta_l^2.
$$
If the minimum is searched over matrices $\mat{A}$ of rank $k=1$, where the matrix will be of the form $\mat{A} = \tilde{\vect{u}}\tilde{\vect{v}}\transp$ (because all columns are multiples of one of the columns) and $\tilde{\vect{u}}$, $\tilde{\vect{v}}$ are non-zero vectors (non necessarily normed, hence the tilde notation), we obtain
$$
\underset{\tilde{\vect{u}},\tilde{\vect{v}}}{\min}\left\|\mat{M} - \tilde{\vect{u}}\tilde{\vect{v}}\transp\right\|_F^2  =  \left\|\mat{M} - \delta_1\vect{u}_1\vect{v}_1\transp\right\|_F^2 =  \sum_{l=2}^{r}\delta_l^2.
$$
Thus, solving
\begin{equation}\label{argminMuv}
(\tilde{\vect{u}}_1,\tilde{\vect{v}}_1) = \underset{\tilde{\vect{u}},\tilde{\vect{v}}}{\argmin}\left\|\mat{M} - \tilde{\vect{u}}\tilde{\vect{v}}\transp\right\|_F^2
\end{equation}
gives us the first left and right singular vectors $\vect{u}_1 = \tilde{\vect{u}}_1 / \|\tilde{\vect{u}}_1\|_2$ and $\vect{v}_1 = \tilde{\vect{v}}_1 / \|\tilde{\vect{v}}_1\|_2$ of \eqref{SVD}, as well as the first singular value $\delta_1 = \|\tilde{\vect{u}}_1\|_2\cdot\|\tilde{\vect{v}}_1\|_2$. Note that this is also equivalent to solve 
$$
\underset{\|\vect{u}\|_2=1,\tilde{\vect{v}}}{\argmin}\left\|\mat{M} - \vect{u}\tilde{\vect{v}}\transp\right\|_F^2\qquad\left(\text{resp. }\underset{\tilde{\vect{u}},\|\vect{v}\|_2=1}{\argmin}\left\|\mat{M} - \tilde{\vect{u}}\vect{v}\transp\right\|_F^2\right)
$$
followed by norming $\tilde{\vect{v}}$ (resp. $\tilde{\vect{u}}$).

\subsection{Penalised SVD}

Shen and Huang \cite{Shen2008} connected expression \eqref{argminMuv} to least squares minimisation in linear regression:
\begin{eqnarray*}
\left\|\mat{M} - \mat{u}\tilde{\mat{v}}\transp\right\|_F^2 & = & \left\|\texttt{vec}(\mat{M}) - (\mat{I}_p \otimes \tilde{\vect{u}})\tilde{\vect{v}}\right\|_2^2\\
& = & \left\|\texttt{vec}(\mat{M}) - (\mat{I}_q \otimes \tilde{\vect{v}})\tilde{\vect{u}}\right\|_2^2.
\end{eqnarray*}

They present a method for sparse principal components by penalising the SVD as follows:
\begin{equation*}
\underset{\|\vect{u}\|_2=1,\tilde{\vect{v}}}{\argmin} \left\|\mat{M} - \mat{u}\tilde{\mat{v}}\transp\right\|_F^2 +P_{\lambda}(\tilde{\vect{v}}), 
\end{equation*}
where $\left\|\mat{M} - \mat{u}\tilde{\mat{v}}\transp\right\|_F^2 = \sum_{i=1}^p\sum_{j=1}^q (m_{ij} - u_i\tilde{v}_j)^2$ is the expanded Frobinus norm, $P_{\lambda}(\tilde{\vect{v}})$ is a penalty function and $\lambda \geq 0$ is a tuning parameter. After solving this problem, they calculate $\vect{v} = \tilde{\vect{v}}/\|\tilde{\vect{v}}\|_2$. Various forms for the penalisation term $P_{\lambda}$ allow for different penalised variable selection techniques.

Following their idea, a number of PLS methods have been proposed based on an iterative algorithm. This algorithm has the basic form:
\begin{enumerate}
\item[$\rhd$] Initialise $\vect{u}$ and $\vect{v}$ to have norm $\|\vect{u}\| = \|\vect{v}\| = 1$;
\item[$\rhd$] Solve
$$\underset{\tilde{\vect{v}}}{\argmin}\left\|\mat{M} - \vect{u}\tilde{\vect{v}}\transp\right\|_F^2 + P_{\lambda_1}(\tilde{\vect{v}});$$
\item[$\rhd$] Normalise $\tilde{\vect{v}}$ to obtain $\vect{v} = \tilde{\vect{v}}/\|\tilde{\vect{v}}\|$;
\item[$\rhd$] Solve
$$\underset{\tilde{\vect{u}}}{\argmin}\left\|\mat{M} - \tilde{\vect{u}}\vect{v}\transp\right\|_F^2 + P_{\lambda_2}(\tilde{\vect{u}});$$
\item[$\rhd$] Normalise $\tilde{\vect{u}}$ to obtain $\vect{u} = \tilde{\vect{u}}/\|\tilde{\vect{u}}\|$;
\end{enumerate}
where the penalty functions $P_{\lambda_1}$ and $P_{\lambda_2}$ enable us to obtain various sparse versions of the SVD. Applying this sparse SVD algorithm to the four standard PLS methods (i)--(iv) gives sparse PLS versions. 

\subsection{Linking SVD to covariance and correlation}\label{linkSVDtocovcor}

It is worthwhile recalling the close connection between SVD and maximum covariance (resp. maximum correlation) analyses; see Appendices~\ref{C1Proof} and \ref{C2Proof}. 
\subsubsection*{(C1)}
The values $\vect{u}_h$ and $\vect{v}_h$ ($h=1,\ldots,r$) in \eqref{SVD} with $\mat{M}=\mat{X}\transp\mat{Y}$ 
solve the minimisation problem 
\begin{eqnarray*}
(\vect{u}_h, \vect{v}_h) &= \underset{\|\vect{u}\|_2^{} = \|\vect{v}\|_2^{} = 1, ~ \delta > 0}{\text{\argmin}}~\|\mat{M} - \delta\vect{u}\vect{v}\transp\|_F^2\\
&= \underset{\|\vect{u}\|_2^{} = \|\vect{v}\|_2^{} = 1}{\argmax}~Cov(\mat{X}\vect{u}, \mat{Y}\vect{v})
\end{eqnarray*}
subject to $\vect{u}\transp\vect{u}_j = \vect{v}\transp\vect{v}_j = 0$, $1\leq j<h$. We note that the vectors are unique up to changes in sign.

Note that the solution to this constrained optimization automatically satisfies:
$$Cov(\mat{X}\vect{u}_h,\mat{Y}\vect{v}_j) = \vect{u}_h\transp\mat{X}\transp\mat{Y}\vect{v}_{j}=0,\qquad 1\leq j<h$$
because the following matrix is diagonal
$$\mat{U}\transp\mat{M}\mat{V} = \mat{U}\transp\mat{U}\mat{\Delta}\mat{V}\transp\mat{V} = \mat{\Delta}.$$
\subsubsection*{(C2)} Suppose that $\mat{X}\transp\mat{X}$ and $\mat{Y}\transp\mat{Y}$ are invertible. The solution to
\begin{eqnarray*}
(\tilde{\vect{w}}_h,\tilde{\vect{z}}_h) & = & \underset{\tilde{\vect{w}},\tilde{\vect{z}}}{\argmax}~Cor(\mat{X}\tilde{\vect{w}},\mat{Y}\tilde{\vect{z}}),\qquad h=1,\ldots,r,
\end{eqnarray*}
subject to the constraints $Cov(\mat{X}\tilde{\vect{w}},\mat{X}\tilde{\vect{w}}_j) = Cov(\mat{Y}\tilde{\vect{z}},\mat{Y}\tilde{\vect{z}}_j) = 0$, $1\leq j<h$ is given by $\tilde{\vect{w}}_h=(\mat{X}\transp\mat{X})^{-1/2}\vect{u}_h$ and $\tilde{\vect{z}}_h = (\mat{Y}\transp\mat{Y})^{-1/2}\vect{v}_h$, where the $\vect{u}_h$ and $\vect{v}_h$ are found through \eqref{SVD} applied with $\mat{M} = (\mat{X}\transp\mat{X})^{-1/2}\mat{X}\transp\mat{Y}(\mat{Y}\transp\mat{Y})^{-1/2}$.
Note that the $\tilde{\vect{w}}_h$'s (resp. the $\tilde{\vect{z}}_h$'s) are not necessarily orthonormal.

\subsection{The four standard PLS methods}\label{fourstandardPLS}

In this section, we survey the four standard PLS methods (i)--(iv) introduced in Section~\ref{Introduction}. At its core, the four PLS methods are used to construct, iteratively, a small number $H\leq r$ (chosen in practice using cross validation techniques) of meaningful linear combinations $\vect{\xi}_h=\mat{X}\vect{w}_h$ and $\vect{\omega}_h=\mat{Y}\vect{z}_h$ (or $\vect{\xi}_h=\mat{X}\tilde{\vect{w}}_h$ and $\vect{\omega}_h=\mat{Y}\tilde{\vect{z}}_h$ for un-normed weights) of the original $X$- and $Y$-variables, with either maximal covariance or correlation. These linear combinations are called component scores, or latent variables. Without additional constraints on the successive scores, there is only one solution for all methods, which is given by the first pair of singular vectors in either \textit{(C1)} or \textit{(C2)}. So it is worthwhile noting that the various PLS methods impose additional \textit{orthogonality} constraints on the optimisation, thus leading to the construction of multiple sets of component scores. Computationally, rather than finding component scores in terms of the original data with the required orthogonality, the PLS algorithms \textit{deflate} the data matrices to ensure that solutions will have the required orthogonality. Component scores are then calculated using the modified (deflated) matrices, and are thus expressed at the $h$-th iteration as $\vect{\xi}_h = \mat{X}_{h-1}\vect{u}_h$ and $\vect{\omega}_h  = \mat{Y}_{h-1}\vect{v}_h$ where $\mat{X}_{h-1}$ and $\mat{Y}_{h-1}$ are the deflated matrices. 

The (normed) weights $\vect{u}_h$ and $\vect{v}_h$ are called the weight vectors (or direction vectors, or saliences, or effective loading weight vectors), while $\vect{w}_h$ and $\vect{z}_h$ (or $\tilde{\vect{w}}_h$ and $\tilde{\vect{z}}_h$ for un-normed versions) are called the adjusted weights. Since the adjusted weights define the score vectors in terms of the original data matrices (as opposed to the deflated matrices), the size of the elements of the weight vector can be interpreted as the effect of the corresponding variables in the component score. On the other hand, the weight vectors $\vect{u}_h$ and $\vect{v}_h$ are defined in terms of the deflated matrices and cannot be interpreted this way. 

The PLS algorithms can be seen as iterative methods that calculate quantities recursively using a deflation step to ensure appropriate orthogonality constraints. The construction of the components leads to decompositions of the original matrices $\mat{X}$ and $\mat{Y}$ of the form:
\begin{equation}\label{decomposition}
\mat{X} =  \mat{\Xi}_{H}\mat{C}_{H}\transp + \mat{F}_{H}^{X},\qquad\mat{Y} = \mat{\Omega}_{H}\mat{D}_{H}\transp + \mat{F}_{H}^{Y},
\end{equation}
where $\mat{\Xi}_H=(\vect{\xi}_j)$ and $\mat{\Omega}_H=(\vect{\omega}_j)$ are called the $X$- and $Y$-scores, $\mat{C}_{H}$ and $\mat{D}_{H}$ are the $X$- and $Y$-loadings, and $\mat{F}_{H}^{X}$ and $\mat{F}_{H}^{Y}$ are the residual matrices.

We now detail the four classical cases (i)--(iv). We state the relevant PLS objective functions for the weight vectors $\vect{u}_h$ and $\vect{v}_h$ at each step $h$, $h=1,\ldots,H$. We describe the deflation method in terms of deflating the matrices $\mat{X}$ and $\mat{Y}$ individually or deflating the matrix $\mat{M} = \mat{X}\transp\mat{Y}$ directly, and the resulting orthogonality. We explicit all terms in the decomposition model~\eqref{decomposition}. The relationship between the weight vectors $\vect{u}_h$ and $\vect{v}_h$ and the adjusted weights $\vect{w}_h$ and $\vect{z}_h$, is given, as well as the PLS objective problem solved by the adjusted weights.

\begin{itemize}
\item[(i)] For PLS-SVD, the roles of $\mat{X}$ and $\mat{Y}$ are symmetric and the analysis focuses on modeling shared information (rather than prediction) as measured by the cross-product matrix $\mat{R}=\mat{X}\transp\mat{Y}$. Note that $\mat{R}$ contains, up to some constant factor, the empirical covariances (resp. correlations) between $X$- and $Y$-variables when the columns of $\mat{X}$ and $\mat{Y}$ are centered (resp. standardised, in which case this method is sometimes called PLSC, for Partial Least Squares Correlation \cite{Krishnan2011}).

The PLS-SVD objective function at step $h$ is given by 
\begin{eqnarray*}
(\vect{w}_h,\vect{z}_h) & = & \underset{\|\vect{w}\|_2^{}=\|\vect{z}\|_2^{}=1}{\argmax}~Cov(\mat{X}\vect{w},\mat{Y}\vect{z}),
\end{eqnarray*}
subject to the constraints $\vect{w}\transp\vect{w}_j=\vect{z}\transp\vect{z}_j=0$, $1\leq j<h$. 
PLS-SVD searches for orthonormal directions $\vect{w}_h$  and orthonormal directions $\vect{z}_h$ ($h=1,\ldots,H$), such that the score vectors $\vect{\xi}_h=\mat{X}\vect{w}_h$ and $\vect{\omega}_h=\mat{Y}\vect{z}_h$ have maximal covariance. Note that the scaling factor $(n-1)^{-1}$ is omited from the covariance (see Subsection~\ref{Notation}) and this has no impact on the $\argmax$ solution. Using (C1), the solutions to this problem are the $H$ first columns of the matrices $\mat{U}$ and $\mat{V}$, which are respectively the left and right singular vectors of $\mat{M}_0 := (\mat{X}\transp\mat{Y})_{0}^{} := \mat{X}\transp\mat{Y}$; see~\eqref{SVD}. 
Another approach is to define $\vect{u}_h=\vect{w}_h$, $\vect{v}_h=\vect{z}_h$, $\mat{X}_0=\mat{X}$, $\mat{Y}_0=\mat{Y}$ and the deflated matrices $\mat{X}_h = \mat{X}_{h-1}(\mat{I} - \vect{u}_h\vect{u}_h\transp) = \mat{X}_0\prod_{j=1}^h(\mat{I}-\vect{u}_j\vect{u}_j\transp)$ and $\mat{Y}_h = \mat{Y}_{h-1}(\mat{I} - \vect{v}_h\vect{v}_h\transp) = \mat{Y}_0\prod_{j=1}^h(\mat{I}-\vect{v}_j\vect{v}_j\transp)$. We have $\vect{u}_h\transp\mat{X}_{h-1}\transp\mat{Y}_{h-1}\vect{v}_h  =  \vect{w}_h\transp\mat{X}\transp\mat{Y}\vect{z}_h$. It is thus possible to replace the objective function with
\begin{eqnarray*}
(\vect{u}_h,\vect{v}_h) & = & \underset{\|\vect{u}\|_2^{}=\|\vect{v}\|_2^{}=1}{\argmax}~Cov(\mat{X}_{h-1}\vect{u},\mat{Y}_{h-1}\vect{v})
\end{eqnarray*}
and compute the previous scores as $\vect{\xi}_h = \mat{X}_{h-1}\vect{u}_h$ and $\vect{\omega}_h = \mat{Y}_{h-1}\vect{v}_h$.


From (C1), and since $\mat{X}$ and $\mat{Y}$ are column-centered,
$$Cov(\vect{\xi}_h,\vect{\omega}_j) 
= 0,\qquad j\ne h.$$
Note that the $X$- (resp. $Y$-) latent variables are not necessarily mutually orthogonal.

Now, because of the orthogonality properties on the $\vect{u}_h$ and $\vect{v}_h$, we have
\begin{eqnarray*}
\vect{u}_{h}\transp\mat{X}\transp\mat{Y}\vect{v}_{h}  & = & \vect{u}_{h}\transp\left(\mat{X}\transp\mat{Y} - \sum_{l=1}^{h-1}\delta_l\vect{u}_l\vect{v}_l\transp\right)\vect{v}_{h}\\
& = & \vect{u}_{h}\transp\mat{M}_{h-1}\vect{v}_{h},
\end{eqnarray*}
where we define $\mat{M}_h = \mat{X}\transp\mat{Y} - \sum_{l=1}^{h}\delta_l\vect{u}_l\vect{v}_l\transp  = \mat{X}_h\transp\mat{Y}_h$. It is thus possible to replace the previous optimization problem with
\begin{eqnarray*}
(\vect{u}_h,\vect{v}_h) & = & \underset{\|\vect{u}\|_2^{}=\|\vect{v}\|_2^{}=1}{\argmax}~\vect{u}\mat{M}_{h-1}^{}\vect{v}.
\end{eqnarray*}
The previous constraints are now automatically satisfied. 
Iterations (deflations) can be done using the relation $\mat{M}_h  =  \mat{M}_{h-1} - \delta_h\vect{u}_h\vect{v}_h\transp$. Thanks to the deflation property of the SVD, we have now that $\delta_h$ is (resp. $\vect{u}_h$ and $\vect{v}_h$ are) the \textit{first} singular value (resp. normed singular vectors) of $\mat{M}_{h-1}$. 

Now, let $\mat{\Xi}_{H}=\mat{X}\mat{U}_{\bullet H}$ and $\mat{\Omega}_{H}=\mat{Y}\mat{V}_{\bullet H}$. 
The decomposition model~\eqref{decomposition} is
\begin{align*}
\mat{X} &= \mathcal{P}_{\mat{\Xi}_{H}}\mat{X} + \mathcal{P}_{\mat{\Xi}_{H}^{\perp}}\mat{X}\\
&= \mat{\Xi}_{H}\mat{\Xi}_{H}^+\mat{X} + \mathcal{P}_{\mat{\Xi}_{H}^{\perp}}\mat{X}\\
&= \mat{\Xi}_{H}\mat{C}_{H}\transp + \mat{F}_{H}^X
\end{align*}
and 
\begin{align*}
\mat{Y} &= \mathcal{P}_{\mat{\Omega}_{H}}\mat{Y} + \mathcal{P}_{\mat{\Omega}_{H}^{\perp}}\mat{Y}\\
&= \mat{\Omega}_{H}\mat{\Omega}_{H}^+\mat{Y} + \mathcal{P}_{\mat{\Omega}_{H}^{\perp}}\mat{Y}\\
&= \mat{\Omega}_{H}\mat{D}_{H}\transp + \mat{F}_{H}^Y,
\end{align*}
with $\mat{C}_{H} = (\mat{\Xi}_{H}^+\mat{X})\transp$ and $\mat{D}_{H} = (\mat{\Omega}_{H}^+\mat{Y})\transp$.\\

\item[(ii)] 
For PLS-W2A, the optimisation problem at step $h$ is
\begin{eqnarray*}
(\vect{u}_h, \vect{v}_h) &= \underset{\|\vect{u}\|_2^{}=\|\vect{v}\|_2^{}=1}{\argmax}~ &Cov(\mat{X}_{h-1}\vect{u},\mat{Y}_{h-1}\vect{v})
\end{eqnarray*}
where the deflated versions of the $\mat{X}$ and $\mat{Y}$ matrices are defined by $\mat{X}_0:=\mat{X}$, $\mat{Y}_0:=\mat{Y}$,
\begin{eqnarray*}
\mat{X}_h & := & \mathcal{P}_{\vect{\xi}_{h}^{\perp}}\mat{X}_{h-1} = \left(\prod_{j=h}^{1}\mathcal{P}_{\vect{\xi}_j^\perp}\right)\mat{X}\\
& = & \left[\mat{I} - \vect{\xi}_{h}(\vect{\xi}_{h}\transp\vect{\xi}_{h})^{-1}\vect{\xi}_{h}\transp\right]\mat{X}_{h-1}
\end{eqnarray*}
and 
\begin{eqnarray*}
\mat{Y}_h & := & \mathcal{P}_{\vect{\omega}_{h}^{\perp}}\mat{Y}_{h-1} \\
& = & \left[\mat{I} - \vect{\omega}_{h}(\vect{\omega}_{h}\transp\vect{\omega}_{h})^{-1}\vect{\omega}_{h}\transp\right]\mat{Y}_{h-1},
\end{eqnarray*}
and where $\vect{\xi}_h=\mat{X}_{h-1}\vect{u}_h$ and $\vect{\omega}_h=\mat{Y}_{h-1}\vect{v}_h$. These score vectors are stored in the matrices $\mat{\Xi}_{H}=(\vect{\xi}_j)$ and $\mat{\Omega}_{H}=(\vect{\omega}_j)$.\\
We have $\mat{X}_ h = (\mat{I} - \mathcal{P}_{\vect{\xi}_h})\mat{X}_{h-1} = (\mat{I} - \mathcal{P}_{\vect{\xi}_h})(\mat{I} - \mathcal{P}_{\vect{\xi}_{h-1}})\mat{X}_{h-2} = (\mat{I} - \mathcal{P}_{\vect{\xi}_{h-1}:\vect{\xi}_h})\mat{X}_{h-2}$ using \cite[Theorem~7, p.~151]{Puntanen2011}. Pursuing this argument leads to 
$\mat{X}_h=\mathcal{P}_{\mat{\Xi}_{\bullet h}^\perp}\mat{X}$, and similarly $\mat{Y}_h=\mathcal{P}_{\mat{\Omega}_{\bullet h}^\perp}\mat{Y}$.
Now $\vect{\xi}_{h}=\mat{X}_{h-1}\vect{u}_{h} = \mathcal{P}_{\mat{\Xi}_{\bullet h-1}^\perp}\mat{X}\vect{u}_{h}$ is clearly orthogonal to $\vect{\xi}_j$ for $j=1,\ldots,h-1$. 
This argument clearly shows that
$$Cov(\vect{\xi}_h,\vect{\xi}_{j}) = Cov(\vect{\omega}_h,\vect{\omega}_{j}) = 0,\qquad 1\leq j<h.$$

PLS-W2A thus searches for successive $X$-score vectors (resp. $Y$-score vectors) that are orthogonal to the previous ones. The first pair $(\vect{\xi}_1,\vect{\omega}_1)$ of $X$- and $Y$- score vectors is the one with maximal covariance. The next pairs are searched for using successively deflated (i.e., after removing the information contained in the previous pairs of scores) versions of $\mat{X}_0$ and $\mat{Y}_0$. We can always write
\begin{equation*}
\mat{X} = \mathcal{P}_{\mat{\Xi}_{H}}\mat{X} + \mathcal{P}_{\mat{\Xi}_{H}^{\perp}}\mat{X}\text{ and }\mat{Y} = \mathcal{P}_{\mat{\Omega}_{H}}\mat{Y} + \mathcal{P}_{\mat{\Omega}_{H}^{\perp}}\mat{Y}.
\end{equation*}
Thus, the elements of the decomposition model \eqref{decomposition} are 
$$\mat{C}_{H} = \mat{X}\transp\mat{\Xi}_{H}(\mat{\Xi}_{H}\transp\mat{\Xi}_{H})^{-1},~~\mat{F}_H^X = \mat{X}_H$$
$$\mat{D}_{H} = \mat{Y}\transp\mat{\Omega}_{H}(\mat{\Omega}_{H}\transp\mat{\Omega}_{H})^{-1},~~\mat{F}_{H}^Y = \mat{Y}_H.$$
At each step, $d_h\vect{u}_h\vect{v}_h\transp$ is the best rank one approximation of $\mat{M}_{h-1}:=\mat{X}_{h-1}\transp\mat{Y}_{h-1}$ in the least squares sense and $\vect{u}_h$ (resp. $\vect{v}_h$) is given by the first left (resp. right) singular vector given by the SVD of $\mat{M}_{h-1}$, where $d_h=Cov(\mat{X}_{h-1}\vect{u}_h,\mat{Y}_{h-1}\vect{v}_h)$ is the first (largest) singular value of this SVD. \\

We have the interesting recursion
\begin{eqnarray*}
\mat{M}_h & = & \mat{X}_h\transp\mat{Y}_h \\
& = & (\mat{X}_{h-1} - \vect{\xi}_h\vect{c}_h\transp)\transp(\mat{Y}_{h-1} - \vect{\omega}_h\vect{e}_h\transp) \\
& = & \mat{X}_{h-1}\transp\mat{Y}_{h-1} - \mat{X}_{h-1}\transp\vect{\omega}_h\vect{e}_h\transp - \vect{c}_h\vect{\xi}_h\transp\mat{Y}_{h-1}\\
&& + \vect{c}_h\vect{\xi}_h\transp\vect{\omega}_h\vect{e}_h\transp\\
& = & \mat{M}_{h-1} - \mat{M}_{h-1}\vect{v}_h\vect{e}_h\transp - \vect{c}_h\vect{u}_h\transp\mat{M}_{h-1}\\
&& + \vect{c}_h\vect{u}_h\transp\mat{M}_{h-1}\vect{v}_h\vect{e}_h\transp\\
& = & (\vect{c}_h\vect{u}_h\transp - \mat{I})\mat{M}_{h-1}(\vect{v}_h\vect{e}_h\transp - \mat{I}).
\end{eqnarray*}

Note that due to the constraints on the $\vect{\xi}_j$, we have that $\mat{\Xi}_{H}\transp\mat{\Xi}_{H}$ is an invertible diagonal matrix and also that $\vect{\xi}_h\transp=\vect{\xi}_h\transp\left(\prod_{j=h-1}^{1}\mathcal{P}_{\vect{\xi}_j^\perp}\right)=\vect{\xi}_h\transp\mathcal{P}_{\mat{\Xi}_{\bullet h-1}^\perp}$. This allows us to write
\begin{eqnarray*}
\mat{C}_{H}\transp & = & (\mat{\Xi}_{H}\transp\mat{\Xi}_{H})^{-1}\mat{\Xi}_{H}\transp\mat{X} \\
& = & (\mat{\Xi}_{H}\transp\mat{\Xi}_{H})^{-1}\left[
\begin{array}{c}
\vect{\xi}_1\transp\\
\vect{\xi}_2\transp\mathcal{P}_{\mat{\Xi}_{\bullet 1}^\perp}\\
\vdots\\
\vect{\xi}_H\transp\mathcal{P}_{\mat{\Xi}_{H-1}^\perp}\\
\end{array}
\right]\mat{X}\\
& = & (\mat{\Xi}_{H}\transp\mat{\Xi}_{H})^{-1}\left[
\begin{array}{c}
\vect{\xi}_1\transp\mat{X}_0\\
\vect{\xi}_2\transp\mat{X}_1\\
\vdots\\
\vect{\xi}_H\transp\mat{X}_{H-1}\\
\end{array}
\right]\\
& = & \left[
\begin{array}{c}
(\vect{\xi}_1\transp\vect{\xi}_1)^{-1}\vect{\xi}_1\transp\mat{X}_0\\
(\vect{\xi}_2\transp\vect{\xi}_2)^{-1}\vect{\xi}_2\transp\mat{X}_1\\
\vdots\\
(\vect{\xi}_H\transp\vect{\xi}_H)^{-1}\vect{\xi}_H\transp\mat{X}_{H-1}\\
\end{array}
\right].
\end{eqnarray*}
Similarly
\begin{eqnarray*}
\mat{D}_{H}\transp &  = & \left[
\begin{array}{c}
(\vect{\omega}_1\transp\vect{\omega}_1)^{-1}\vect{\omega}_1\transp\mat{Y}_0\\
(\vect{\omega}_2\transp\vect{\omega}_2)^{-1}\vect{\omega}_2\transp\mat{Y}_1\\
\vdots\\
(\vect{\omega}_H\transp\vect{\omega}_H)^{-1}\vect{\omega}_H\transp\mat{Y}_{H-1}\\
\end{array}
\right].
\end{eqnarray*}
These are the expressions given e.g., in \cite[p.~10]{Wegelin2000}. \\

The linear combinations $\vect{\xi}_h = \mat{X}_{h-1}\vect{u}_h$, and $\vect{\omega}_h = \mat{Y}_{h-1}\vect{v}_h$ are searched for recursively in the space spanned by the previous residuals. In what follows, we will consider how these linear combinations can be expressed in terms of the original variables. From Appendix~\ref{DecompositionPLSR}, we can write $\mat{X}_h = \mat{X}\mat{A}^{(h)}$ and $\mat{Y}_h = \mat{Y}\mat{B}^{(h)}$ with 
$$\mat{A}^{(h)} = \prod_{j=1}^h(\mat{I} - \vect{u}_j(\vect{\xi}_j\transp\vect{\xi}_j)^{-1}\vect{\xi}_j\transp\mat{X})$$
and
$$\mat{B}^{(h)} = \prod_{j=1}^h(\mat{I} - \vect{v}_j(\vect{\omega}_j\transp\vect{\omega}_j)^{-1}\vect{\omega}_j\transp\mat{Y}).$$

Defining $\tilde{\vect{w}}_h = \prod_{j=1}^{h-1}(\mat{I} - \vect{u}_j(\vect{\xi}_j\transp\vect{\xi}_j)^{-1}\vect{\xi}_j\transp\mat{X})\vect{u}_h$ and $\tilde{\vect{z}}_h = \prod_{j=1}^{h-1}(\mat{I} - \vect{v}_j(\vect{\omega}_j\transp\vect{\omega}_j)^{-1}\vect{\omega}_j\transp\mat{Y})\vect{v}_h$, we have that $\vect{\xi}_h = \mat{X}\tilde{\vect{w}}_h$, and $\vect{\omega}_h = \mat{Y}\tilde{\vect{z}}_h.$ These adjusted weights describe the effect of each of the original variables in constructing the scores $\vect{\xi}_h$ and $\vect{\omega}_h$. To find what objective function these weights solve, we can use the argument from Appendix~\ref{AdjustedWeightProblem} to find:
$$
\vect{u}_h = \mathcal{P}_{\tilde{\mat{W}}_{\bullet h-1}^\perp}\tilde{\vect{w}}_h,\qquad
\vect{v}_h = \mathcal{P}_{\tilde{\mat{Z}}_{\bullet h-1}^\perp}\tilde{\vect{z}}_h.
$$

Substituting these equations into the objective function for the $h$-th set of PLS-W2A adjusted weights $(\tilde{\vect{w}}_h, \tilde{\vect{z}}_h)$, gives the equivalent optimisation problem
\begin{eqnarray*}\label{W2A_PLS_OV}
\underset{\|\mathcal{P}_{\tilde{\mat{W}}_{\bullet h-1}^\perp}\tilde{\vect{w}}\|_2^{}=\|\mathcal{P}_{\tilde{\mat{Z}}_{\bullet h-1}^\perp}\tilde{\vect{z}}\|_2^{}=1}{\text{argmax}}~ &Cov(\mat{X}\tilde{\vect{w}},\mat{Y}\tilde{\vect{z}}).
\end{eqnarray*}

\item[(iii)] %
The CCA objective function at step $h$ is given by
$$(\tilde{\vect{w}}_h,\tilde{\vect{z}}_h)=\underset{\tilde{\vect{w}},\tilde{\vect{z}}}{\argmax}~Cor(\mat{X}\tilde{\vect{w}},\mat{Y}\tilde{\vect{z}}),$$
subject to the constraints
$$Cov(\mat{X}\tilde{\vect{w}},\mat{X}\tilde{\vect{w}}_{j}) = Cov(\mat{Y}\tilde{\vect{z}},\mat{Y}\tilde{\vect{z}}_{j}) = 0,~1\leq j<h.$$

Classical CCA relates $\mat{X}$ and $\mat{Y}$ by maximising the \textit{correlation} between the scores (or canonical variates) $\vect{\xi}_h=\mat{X}\tilde{\vect{w}}_h$ and $\vect{\omega}_h=\mat{Y}\tilde{\vect{z}}_h$, but without imposing a unit norm to the adjusted weights (or canonical) vectors $\tilde{\vect{w}}_h$ and $\tilde{\vect{z}}_h$. 
\\
From the proof of (C2), and assuming that the $\mat{X}$ and $\mat{Y}$ sample covariance matrices are nonsingular (more on this later), an equivalent CCA objective function at step $h$ is given by
$$\left\{
\begin{array}{c}
(\vect{u}_h,\vect{v}_h)=\underset{\|\vect{u}\|_2^{}=\|\vect{v}\|_2^{}=1}{\argmax}~\vect{u}\transp\mat{M}_{0}^{}\vect{v}, \\
\tilde{\vect{w}}_h=(\mat{X}\transp\mat{X})^{-1/2}\vect{u}_h~\text{ and }~\tilde{\vect{z}}_h = (\mat{Y}\transp\mat{Y})^{-1/2}\vect{v}_h,
\end{array}
\right.$$
subject to the constraints $\vect{u}\transp\vect{u}_j = \vect{v}\transp\vect{v}_j=0$, $1\leq j<h$, with $\mat{M}_0^{}:=(\mat{X}\transp\mat{X})^{-1/2}\mat{X}\transp\mat{Y}(\mat{Y}\transp\mat{Y})^{-1/2}$. \\

Using (C2), the solution $\vect{u}_h$ (resp. $\vect{v}_h$) to this problem is the $h$-th column of the matrix $\mat{U}$ (resp. $\mat{V}$), obtained by applying~\eqref{SVD} to $\mat{M}_0$. 
Now, because of the imposed constraints on the $\vect{u}_h$ and $\vect{v}_h$, we have
\begin{eqnarray*}
\vect{u}_h\transp\mat{M}_0\vect{v}_h & = & \vect{u}_h\transp\left(\mat{M}_0 - \sum_{l=1}^{h-1}\delta_h\vect{u}_l\vect{v}_l\transp\right)\vect{v}_h\\
& := & \vect{u}_h\transp\mat{M}_{h-1}\vect{v}_h.
\end{eqnarray*}
It is thus possible to replace the above objective function
with
$$\left\{
\begin{array}{c}
(\vect{u}_h,\vect{v}_h)=\underset{\|\vect{u}\|_2^{}=\|\vect{v}\|_2^{}=1}{\argmax}~\vect{u}\transp\mat{M}_{h-1}^{}\vect{v}, \\
\tilde{\vect{w}}_h=(\mat{X}\transp\mat{X})^{-1/2}\vect{u}_h~\text{ and }~\tilde{\vect{z}}_h = (\mat{Y}\transp\mat{Y})^{-1/2}\vect{v}_h,
\end{array}
\right.$$
where, thanks to the deflation property of the SVD, we have that 
$\delta_h$, $\vect{u}_h$ and $\vect{v}_h$ are now obtained respectively (and successively) as the \textit{first} singular value and \textit{first} singular vectors of $\mat{M}_{h-1}^{}$. Iterations (deflations) are done using the relation $\mat{M}_h  =  \mat{M}_{h-1} - \delta_h\vect{u}_h\vect{v}_h\transp$. Another approach is to define $\mat{X}_0 = \mat{X}(\mat{X}\transp\mat{X})^{-1/2}$, $\mat{Y}_0 = \mat{Y}(\mat{Y}\transp\mat{Y})^{-1/2}$, $\mat{X}_h = \mat{X}_{h-1} ( \mat{I} - \vect{u}_h \vect{u}_h\transp)$ and $\mat{Y}_h = \mat{Y}_{h-1} ( \mat{I} - \vect{v}_h \vect{v}_h\transp)$. We have $\mat{M}_h = \mat{X}_h\transp\mat{Y}_h$. It follows that%
\begin{eqnarray*}
\mat{X}_h\transp\mat{Y}_h & = & (\mat{I} - \vect{u}_h\vect{u}_h\transp)\mat{X}_{h-1}\transp\mat{Y}_{h-1}(\mat{I} - \vect{v}_{h}\vect{v}_{h}\transp)\\
& = & \prod_{i=h}^1 (\mat{I} - \vect{u}_i\vect{u}_i\transp)\mat{X}_0\transp\mat{Y}_0\prod_{i=1}^h(\mat{I} - \vect{v}_i\vect{v}_i\transp)\\
& = & (\mat{I} - \mat{U}_{\bullet h}\mat{U}_{\bullet h}\transp)\mat{U}\mat{\Delta}\mat{V}\transp(\mat{I} - \mat{V}_{\bullet h}\mat{V}_{\bullet h}\transp)\\
& = & (\mat{U}\mat{\Delta}\mat{V}\transp - \mat{U}_{\bullet h}\mat{\Delta}_h\mat{V}_{\bullet h}\transp)(\mat{I} - \mat{V}_{\bullet h}\mat{V}_{\bullet h}\transp)\\
& = & \mat{U}\mat{\Delta}\mat{V}\transp - \mat{U}\mat{\Delta}\mat{V}\transp\mat{V}_{\bullet h}\mat{V}_{\bullet h}\transp - \mat{U}_{\bullet h}\mat{\Delta}_{h}\mat{V}_{\bullet h}\transp \\
&& 
+ \mat{U}_{\bullet h}\mat{\Delta}_h\mat{V}_{\bullet h}\transp\\
& = & \mat{U}\mat{\Delta}\mat{V}\transp -\mat{U}_{\bullet h}\mat{\Delta}_{h}\mat{V}_{\bullet h}\transp\\
& = & \mat{M}_0 - \sum_{i=1}^h\delta_i\vect{u}_i\vect{v}_i\transp
\end{eqnarray*}
where the product $\prod_{i=1}^h(\mat{I} - \vect{u}_i\vect{u}_i\transp) = (\mat{I} - \mat{U}_{\bullet h}\mat{U}_{\bullet h}\transp)$ follows from \cite[Theorem~7, p.~151]{Puntanen2011}, and where $\mathcal{P}_{\vect{u}_h}=\vect{u}_h\vect{u}_h\transp$.\\

It is thus possible to replace the objective function with
\begin{eqnarray*}
(\vect{u}_h,\vect{v}_h) & = & \underset{\|\vect{u}\|_2^{}=\|\vect{v}\|_2^{}=1}{\argmax}~Cov(\mat{X}_{h-1}\vect{u},\mat{Y}_{h-1}\vect{v}),
\end{eqnarray*}
and to define the scores as $\vect{\xi}_h = \mat{X}_{h-1}\vect{u}_h$ and $\vect{\omega}_h = \mat{Y}_{h-1}\vect{v}_h$.

Note that orthogonality of the scores holds due to the SVD properties:
\begin{align*}
\mat{\xi}_j\transp\vect{\xi}_h &= \tilde{\vect{w}}\transp_j\mat{X}\transp\mat{X}\tilde{\vect{w}}_h\\
&= \vect{u}\transp_j(\mat{X}\transp\mat{X})^{-1/2}\mat{X}\transp\mat{X}(\mat{X}\transp\mat{X})^{-1/2}\vect{u}_h\\
&= \vect{u}\transp_j\vect{u}_h\\
&= 0
\end{align*}
for all $j \neq h.$ Similarly, we find $\vect{\omega}_j\transp\vect{\omega}_h = 0$ for $j \neq h$. We also have orthogonality between $X$- and $Y$-scores. For all $j \neq h$,
\begin{align*}
\mat{\xi}_j\transp\vect{\omega}_h &= \tilde{\vect{w}}\transp_j\mat{X}\transp\mat{Y}\tilde{\vect{z}}_h\\
&= \vect{u}\transp_j(\mat{X}\transp\mat{X})^{-1/2}\mat{X}\transp\mat{Y}(\mat{Y}\transp\mat{Y})^{-1/2}\vect{v}_h\\
&= \vect{u}\transp_j\mat{M}_0\vect{v}_h\\
&= \vect{u}\transp_j\left(\sum_{l = 1}^r\delta_l\vect{u}_l\vect{v}_l\transp\right)\vect{v}_h\\
&= 0.
\end{align*}
\\
Let $\mat{\Xi}_{H} = \mat{X}\tilde{\mat{W}}_{H}$ and $\mat{\Omega}_{H} = \mat{Y}\tilde{\mat{Z}}_{H}$, where $\tilde{\mat{W}}_{H} = (\mat{X}\transp\mat{X})^{-1/2}\mat{U}_{\bullet H}$ and $\tilde{\mat{Z}}_{H} = (\mat{Y}\transp\mat{Y})^{-1/2}\mat{V}_{\bullet H}$. Since we have assumed that $\mat{X}\transp\mat{X}$ is invertible, we have
\begin{align*}
\mat{\Xi}_{H}\transp\mat{\Xi}_{H} &=\mat{U}_{\bullet H}\transp(\mat{X}\transp\mat{X})^{-1/2}\mat{X}\transp\mat{X}(\mat{X}\transp\mat{X})^{-1/2}\mat{U}_{\bullet H}\\
&=\mat{U}_{\bullet H}\transp\mat{U}_{\bullet H}\\
&= \mat{I}_H.
\end{align*}
Similarly, $\mat{\Omega}_{H}\transp\mat{\Omega}_{H} = \mat{I}_H$. We have
\begin{align*}
\mat{X} &= \mathcal{P}_{\mat{\Xi}_{H}}\mat{X} + \mathcal{P}_{\mat{\Xi}_{H}^{\perp}}\mat{X}\\
&= \mat{\Xi}_{H}(\mat{\Xi}_{H}\transp\mat{\Xi}_{H})^{-1}\mat{\Xi}_{H}\transp\mat{X} + \mathcal{P}_{\mat{\Xi}_{H}^{\perp}}\mat{X}\\
&= \mat{\Xi}_{H}\mat{\Xi}_{H}\transp\mat{X} + \mathcal{P}_{\mat{\Xi}_{H}^{\perp}}\mat{X}.
\end{align*}
A similar expression holds for $\mat{Y}$. Thus the elements of the decomposition model~\eqref{decomposition} are
$$\mat{C}_{H} = \mat{X}\transp\mat{\Xi}_{H},~~\mat{F}_H^X = \mathcal{P}_{\mat{\Xi}_{H}^{\perp}}\mat{X};$$
$$\mat{D}_{H} = \mat{Y}\transp\mat{\Omega}_{H},~~\mat{F}_{H}^Y = \mathcal{P}_{\mat{\Omega}_{H}^{\perp}}\mat{Y}.$$
It has been suggested \cite[p.~287]{Mardia79}, \cite[p.~75]{Nielsen2002} to use generalised inverses (e.g., Moore-Penrose) to deal with the singular case, and use the objective function at step~$h$ 
$$\left\{
\begin{array}{c}
(\vect{u}_h,\vect{v}_h)=\underset{\|\vect{u}\|_2^{}=\|\vect{v}\|_2^{}=1}{\argmax}~\vect{u}\transp\mat{M}_{0}^{}\vect{v}, \\
\tilde{\vect{w}}_h=(\mat{X}\transp\mat{X})^{+1/2}\vect{u}_h~\text{ and }~\tilde{\vect{z}}_h = (\mat{Y}\transp\mat{Y})^{+1/2}\vect{v}_h,
\end{array}
\right.$$
where $\mat{M}_0 = (\mat{X}\transp\mat{X})^{+1/2}\mat{X}\transp\mat{Y}(\mat{Y}\transp\mat{Y})^{+1/2}$.


This being said, this approach can produce a meaningless solution, with correlations trivially equal to one. Indeed, there exists infinitely many other generalised inverses. They might lead to other weights and scores, still with the same optimal correlation between scores. Moreover, in this case, a small change in the data can lead to large changes in the weights and scores \cite[pp.~26--27]{Wegelin2000}. In other words, overfitting would occur.

An alternative for the case of singular matrices is to perform regularisation on the sample covariance matrices. The regularised solution trades off bias for a lower variance solution. Regularisation was first introduced to the CCA method by \cite{Vinod1976} and later refined by \cite{Leurgans1993}. This method is known as regularised CCA (rCCA) or canonical ridge analysis and is closely related to Tikhonov's regularisation (or ridge regression) for the solution of systems of linear equations. This regularisation is imposed by replacing the matrices $\mat{X}\transp\mat{X}$ and $\mat{Y}\transp\mat{Y}$ with $\mat{X}\transp\mat{X} + \lambda_x\mat{I}_p$ and $\mat{Y}\transp\mat{Y}+\lambda_y\mat{I}_q$ respectively in the optimisation criterion. The regularisation parameters $\lambda_x$ and $\lambda_y$ should be nonnegative and if they are nonzero, then the regularised covariance matrices will be nonsingular. With a slightly different application of the regularisation parameters we could use:
\begin{align*}
&(1-\lambda_x^*)\mat{X}\transp\mat{X} + \lambda_x^*\mat{I}_{p}\\
&(1-\lambda_y^*)\mat{Y}\transp\mat{Y} + \lambda_y^*\mat{I}_{q},
\end{align*}
with $0\leq \lambda_x^*, \lambda_y^* \leq 1$. The objective function in this case changes to \cite[p.~38]{Rosipal2006}:
$$\left\{
\begin{array}{c}
(\vect{u}_h,\vect{v}_h)=\underset{\|\vect{u}\|_2^{}=\|\vect{v}\|_2^{}=1}{\argmax}~\vect{u}\transp\mat{M}_{0}^{}\vect{v}, \\
\tilde{\vect{w}}_h=((1-\lambda_x^*)\mat{X}\transp\mat{X} + \lambda_x^*\mat{I}_{p})^{-1/2}\vect{u}_h\\
\tilde{\vect{z}}_h = ((1-\lambda_y^*)\mat{Y}\transp\mat{Y} + \lambda_y^*\mat{I}_{q})^{-1/2}\vect{v}_h,
\end{array}
\right.
$$
where $\mat{M}_0$ is defined as
$$((1-\lambda_x^*)\mat{X}\transp\mat{X} + \lambda_x^*\mat{I}_{p})^{-1/2}\mat{X}\transp\mat{Y}((1-\lambda_y^*)\mat{Y}\transp\mat{Y} + \lambda_y^*\mat{I}_{q})^{-1/2}.$$
Note that ordinary CCA is obtained at $\lambda_x^* = \lambda_y^* = 0$, and PLS-SVD is obtained with $\lambda_x^* = \lambda_y^* = 1$. Other approaches exist; see e.g., \cite[eq.~(13)]{Witten2009}.

\item[(iv)] PLS-R (also called PLS1 if $q=1$ or PLS2 if $q>1$) is a regression technique that predicts one set of data from another, hence termed asymmetric, while describing their common structure. It finds latent variables (also called component scores) that model $\mat{X}$ and simultaneously predict $\mat{Y}$. While several algorithms have been developed to solve this problem, we focus on the two most well known variants. The first, is an extension of the Nonlinear estimation by Iterative PArtial Least Squares (NIPALS), initially proposed by H. Wold \cite{Wold1966} as an alternative algorithm for  implementing  Principal  Component  Analysis, and modified by \cite{Wold1984} to obtain a regularized component based regression tool. The second, is the Statistically Inspired Modification of PLS (SIMPLS) \cite{deJong1993}. We now give some details about outputs of these two algorithms. Other PLS regression algorithms can be found in \cite{Lindgren1998}, and see also \cite{Alin2009} for a numerical comparison.\\

The $h$-th set of PLS regression weights $(\vect{u}_h, \vect{v}_h)$ given by NIPALS solve the optimisation problem \cite[eq. (5)]{Braak1998}
\begin{eqnarray*}
(\vect{u}_h, \vect{v}_h) &= \underset{\|\vect{u}\|_2^{} = \|\vect{v}\|_2^{}=1}{\argmax}~Cov(\mat{X}_{h-1}\vect{u},\mat{Y}_{h-1}\vect{v}),
\end{eqnarray*}
where the deflated matrices are defined by $\mat{X}_0:=\mat{X}$, $\mat{Y}_0:=\mat{Y}$, 
\begin{align*}
\mat{X}_h &= \mat{X}_{h-1} - \vect{\xi}_h(\vect{\xi}_{h}\transp\vect{\xi}_h)^{-1}\vect{\xi}_h\transp\mat{X}_{h-1} = \mathcal{P}_{\vect{\xi}_{h}^{\perp}}\mat{X}_{h-1},
\end{align*}
with $\vect{\xi}_h=\mat{X}_{h-1}\vect{u}_h$,  and
\begin{align*}
\mat{Y}_h &= \mat{Y}_{h-1} - \vect{\xi}_h(\vect{\xi}_h\transp\vect{\xi}_h)^{-1}\vect{\xi}_h\transp\mat{Y}_{h-1} = \mathcal{P}_{\vect{\xi}_{h}^{\perp}}\mat{Y}_{h-1}.
\end{align*}
Classical PLS-R searches for successive $X$-score vectors $\vect{\xi}_h$ (stored in the matrix $\mat{\Xi}_H$) that are orthogonal to the previous ones by construction ($\vect{\xi}_{h}= \mathcal{P}_{\mat{\Xi}_{\bullet h-1}^\perp}\mat{X}\vect{u}_{h}$)
and $Y$-score vectors $\vect{\omega}_h$ (defined below and stored in the matrix $\mat{\Omega}_H$). The
first pair $(\vect{\xi}_1,\vect{\omega}_1)$ of $X$- and $Y$-score vectors is the one
with maximal covariance. 
The next pairs are searched
for using successively deflated versions of $\mat{X}$ and of $\mat{Y}$. We thus remove the
information contained in the previous $X$-scores here. From (C1), the solution $\vect{u}_h$ (resp. $\vect{v}_h$) is the \textit{first} left (resp. right) singular vector of $\mat{M}_{h-1}:=\mat{X}_{h-1}\transp\mat{Y}_{h-1}$. \\

We have the interesting recursion
\begin{eqnarray*}
\mat{M}_h & = & \mat{X}_h\transp\mat{Y}_h \\
& = & (\mat{X}_{h-1} - \vect{\xi}_h\vect{c}_h\transp)\transp(\mat{Y}_{h-1} - \vect{\xi}_h\vect{d}_h\transp) \\
& = & \mat{X}_{h-1}\transp\mat{Y}_{h-1} - \mat{X}_{h-1}\transp\vect{\xi}_h\vect{d}_h\transp - \vect{c}_h\vect{\xi}_h\transp\mat{Y}_{h-1}\\
&& + \vect{c}_h\vect{\xi}_h\transp\vect{\xi}_h\vect{d}_h\transp\\
& = & \mat{M}_{h-1} - \mat{X}_{h-1}\transp\mat{X}_{h-1}\vect{u}_h\vect{d}_h\transp - \vect{c}_h\vect{u}_h\transp\mat{M}_{h-1}\\
&& + \vect{c}_h\vect{u}_h\transp\mat{X}_{h-1}\transp\mat{X}_{h-1}\vect{u}_h\vect{d}_h\transp\\
& = & (\mat{I} - \vect{c}_h\vect{u}_h\transp)\mat{M}_{h-1} - (\mat{I} - \vect{c}_h\vect{u}_h\transp)\mat{N}_{h-1}\vect{u}_h\vect{d}_h\transp\\
& = & (\mat{I} - \vect{c}_h\vect{u}_h\transp)(\mat{M}_{h-1} - \mat{N}_{h-1}\vect{u}_h\vect{d}_h\transp),
\end{eqnarray*}
where $\mat{N}_0 = \mat{X}_0\transp\mat{X}_0$ and 
\begin{eqnarray*}
\mat{N}_h & = & (\vect{c}_h\vect{u}_h\transp - \mat{I})\mat{N}_{h-1}(\vect{u}_h\vect{c}_h\transp - \mat{I}).
\end{eqnarray*}

Note that the original NIPALS algorithm computes the above quantities using an iterative procedure designed to compute eigenvectors (but see the relation between eigenvectors and singular vectors in Appendix~\ref{linkEigenElements}). Two versions are found in the literature, wether $\vect{v}_h$ is scaled \cite{Hoskuldsson1988} or not \cite{Wold1984, Tenenhaus1998}. At the end of both algorithms, the fitted values $\widehat{\mat{Y}}_H$ are computed \cite[Equ.~(20)]{Phatak1997}
\begin{align*}
\widehat{\mat{Y}}_H &= \mathcal{P}_{\mat{\Xi}_{H}}\mat{Y} = \mat{\Xi}_{H}(\mat{\Xi}_{H}\transp\mat{\Xi}_{H})^{-1}\mat{\Xi}_{H}\transp\mat{Y}.
\end{align*}
This is described in Appendix~\ref{TwoNIPALSversions}.\\

Let $\alpha_h = \|\vect{\xi}_h\|^2 / \|\mat{Y}_{h-1}\transp\vect{\xi}_h\|$. Now, define $p_h=\alpha_h^{-1}$ in the scaled case and $p_h=1$ otherwise. The $Y$-score vectors are defined as $\vect{\omega}_h = p_h\alpha_h\mat{Y}_{h-1}\vect{v}_h$, $h=1,\ldots,H$.\\
In addition to the usual decomposition equations~\eqref{decomposition} that will be explicited below, the PLS regression algorithm includes an additional ``inner relationship'' which relates the $Y$-scores $\mat{\Omega}_{\bullet h}$ to the $X$-scores $\mat{\Xi}_{\bullet h}$ explicitly:
\begin{eqnarray}\label{eq:inner}
\mat{\Omega}_{\bullet h} 
& = & \mat{\Xi}_{\bullet h}\mat{P}_h + \mat{R}_{\bullet h},
\end{eqnarray}
where $\mat{\Omega}_{\bullet h} = (\vect{\omega}_j)_{1\leq j\leq h}$, $\mat{P}_h = \textrm{diag}(p_j)_{1\leq j\leq h}$ and where $\mat{R}_{\bullet h}$ is a matrix of residuals. Note that in the unscaled case, $\mat{P}_h=\mat{I}$. Proof is provided in Appendix~\ref{ProofInner}. \\

The decomposition model~\eqref{decomposition} is given by (see Appendix~\ref{DecompositionPLSR} for proof):
\begin{align*}
\mat{X} &=\mathcal{P}_{\vect{\Xi}_{H}}\mat{X} + \mathcal{P}_{\vect{\Xi}_{H}^{\perp}}\mat{X}
=\mat{\Xi}_{H}\mat{C}_{H}\transp + \mat{X}_{H},
\end{align*}
where $\mat{C}_{H} = \mat{X}\transp\mat{\Xi}_{H}(\mat{\Xi}_{H}\transp\mat{\Xi}_{H})^{-1}$ and where the matrix of residuals is $\mat{F}_H^X = \mat{X}_{H}=\mathcal{P}_{\vect{\Xi}_{H}^{\perp}}\mat{X}$. We have
\begin{eqnarray*}
\mat{Y} &=& \mat{\Omega}_{H}\mat{D}_{H}\transp + \mat{F}_{H}^Y\\
& = & (\mat{\Xi}_{H}\mat{P}_H + \mat{R}_{H})\mat{D}_{H}\transp + \mat{F}_{H}^Y\\
& = & \mat{\Xi}_{H}\mat{P}_H\mat{D}_{H}\transp + (\mat{R}_{H}\mat{D}_{H}\transp + \mat{F}_{H}^Y)\\
& = & \mat{X}\tilde{\mat{W}}_{H}\mat{P}_H\mat{D}_{H}\transp + (\mat{R}_{H}\mat{D}_{H}\transp + \mat{F}_{H}^Y)\\
& = & \mat{X}\widehat{\mat{B}}_{PLS} + \mat{E}_{H}^Y,
\end{eqnarray*}
where $\mat{D}_H=[\vect{v}_1,\ldots,\vect{v}_H]$, $\widehat{\mat{B}}_{PLS} = \mat{U}_{\bullet H}(\mat{C}_{H}\transp\mat{U}_{\bullet H})^{-1}\mat{P}_H\mat{D}_{H}\transp := \tilde{\mat{W}}_{H}\mat{P}_H\mat{D}_{H}\transp$, and where the matrices of residuals are $\mat{F}_{H}^Y = \mat{\Xi}_H\mat{G}_H\transp - \mat{\Omega}_H\mat{D}_H\transp + \mat{Y}_H$ and $\mat{E}_{H}^Y = \mat{R}_{H}\mat{D}_{H}\transp + \mat{F}_{H}^Y$. The $h$-th row of $\mat{C}_{H}\transp$ and $\mat{G}_{H}\transp$ are respectively $(\vect{\xi}_h\transp\vect{\xi}_h)^{-1}\vect{\xi}_h\transp\mat{X}_{h-1}$ and $(\vect{\xi}_h\transp\vect{\xi}_h)^{-1}\vect{\xi}_h\transp\mat{Y}_{h-1}$.\\

\begin{remark}
For univariate regression, the objective function can be restated as follows \cite{Frank1993}:
\begin{eqnarray*}
\vect{u}_h = \underset{\|\vect{u}\|_2^{}=1}{\argmax}~Cor^2(\mat{X}_{h-1}\vect{u},\mat{Y}_{h-1})Var(\mat{X}_{h-1}\vect{u}),
\end{eqnarray*}
where $\vect{v} = 1$ since the response $\mat{Y}_{h-1}:n\times1$ is univariate, and where we have used the relationship
\begin{align*}
Cov^2(&\mat{X}_{h-1}\vect{u},\mat{Y}_{h-1})=\\&Var(\mat{X}_{h-1}\vect{u})Cor^2(\mat{X}_{h-1}\vect{u},\mat{Y}_{h-1})Var(\mat{Y}_{h-1}).
\end{align*} 
This formulation shows that PLS seeks directions that relate $\mat{X}$ and $\mat{Y}$ by maximising the correlation, and capture the most variable directions in the $X$-space. \\
\end{remark}

We now present equivalent objective functions that one can encounter in the literature. Since the optimal solution $\vect{v}_h$ to the objective problem should be proportional to $\mat{M}_{h-1}\transp\vect{u}_h$ (see Proof of (C1) in the Appendix~\ref{C1Proof}), the optimisation problem is equivalent to \cite[eq. (13)--(14)]{Burnham1996}
\begin{eqnarray*}
\left\{
\begin{array}{l}
\vect{u}_h  =  \underset{\|\vect{u}\|_2^{}=1}{\argmax} (\vect{u}\transp\mat{M}_{h-1}\mat{M}_{h-1}\transp\vect{u}) \\
\tilde{\vect{v}}_h = \mat{M}_{h-1}\transp\vect{u}_h ~;~\text{norm }\tilde{\vect{v}}_h,
\end{array}
\right.
\end{eqnarray*}
whose solution can be obtained using the so-called PLS2 algorithm \cite{Braak1998}. 
We note that only one of $\mat{X}$ or $\mat{Y}$ needs to be deflated, \cite{Braak1998} because:
\begin{align*}
\mat{M}_h &= \mat{X}_h\transp\mat{Y}_h\\
&= \mat{X}\transp\mathcal{P}_{\vect{\Xi}_{\bullet h}^{\perp}}\transp\mathcal{P}_{\vect{\Xi}_{\bullet h}^{\perp}}\mat{Y}\\
&= \mat{X}\transp\mathcal{P}_{\vect{\Xi}_{\bullet h}^{\perp}}\transp\mat{Y},
\end{align*}
which is equal to $\mat{X}_h\transp\mat{Y}$ (or $\mat{X}\transp\mat{Y}_h$). Thus the previous optimisation problem can be written as \cite[eq. (7)]{Braak1998}:
\begin{eqnarray*}
\left\{
\begin{array}{l}
\vect{u}_h  =  \underset{\|\vect{u}\|_2^{}=1}{\argmax} (\vect{u}\transp\mat{X}_{h-1}\transp\mat{Y}\mat{Y}\transp\mat{X}_{h-1}\vect{u}) \\
\tilde{\vect{v}}_h = \mat{Y}\transp\mat{X}_{h-1}\vect{u}_h ~;~\text{norm }\tilde{\vect{v}}_h.
\end{array}
\right.
\end{eqnarray*}
Similar to PLS-W2A, the linear combinations $\vect{\xi}_h = \mat{X}_{h-1}\vect{u}_h$ are searched for recursively through the successive residuals. We now consider the construction of the scores in terms of the original variables $\vect{\xi}_h = \mat{X}\tilde{\vect{w}}_h$. From Appendix~\ref{AdjustedWeightProblem}, we have
$$
\vect{u}_h = \mathcal{P}_{\tilde{\mat{W}}_{\bullet h-1}^\perp}\tilde{\vect{w}}_h.
$$
The above optimisation problem is thus equivalent to solving \cite[eq. (2)]{Chun2010}
\begin{eqnarray}\label{NIPALS_PLS_OV}
\left\{
\begin{array}{l}
\tilde{\vect{w}}_h   =   \underset{\|\mathcal{P}_{\tilde{\mat{W}}_{\bullet h-1}^{\perp}}\tilde{\vect{w}}\|_2=1}{\argmax} (\tilde{\vect{w}}\transp\mat{X}\transp\mat{Y}\mat{Y}\transp\mat{X}\tilde{\vect{w}}) \\
\tilde{\vect{v}}_h = \mat{Y}\transp\mat{X}\tilde{\vect{w}}_h ~;~\text{norm }\tilde{\vect{v}}_h,
\end{array}
\right.
\end{eqnarray}
(without deflations), this is the so-called ``PLS2'' objective function.\\

The second most commonly used PLSR algorithm, called SIMPLS \cite{deJong1993}, 
calculates the PLS latent components directly as linear combinations of the original variables. The objective function to optimise is \cite[eq. (3)]{Chun2010}
\begin{eqnarray*}
\vect{w}_h   =   \underset{\|\vect{w}\|_2=1}{\argmax} (\vect{w}\transp\mat{X}\transp\mat{Y}\mat{Y}\transp\mat{X}\vect{w}),
\end{eqnarray*}
(without deflations) subject to the constraints
$$Cov(\mat{X}\vect{w},\mat{X}\vect{w}_j) 
=0, ~ 1\leq j<h,$$
from which we compute
$$
\vect{u}_h   = \vect{w}_h,\qquad 
\vect{v}_h  =  \mat{Y}\transp\mat{X}\vect{w}_h/\|\mat{Y}\transp\mat{X}\vect{w}_h\|_2.
$$

It is important to note that both algorithms have the same objective function but different constraints and thus yield different sets of direction vectors. The solution $\vect{w}_h$ to SIMPLS is the first left singular vector of $\mat{M}_{h-1}:=\mathcal{P}_{\mat{C}_{\bullet h-1}^\perp}\mat{X}\transp\mat{Y}$ \cite[p.~322]{Phatak1997}.\\

\begin{remark}
Another equivalent objective function for SIMPLS is \cite{Boulesteix2007}
$$
(\vect{w}_h,\vect{v}_h) = \underset{\|\vect{w}\|_2^{}=\|\vect{v}\|_2^{}=1}{\argmax}(\vect{w}\transp\mat{X}\transp\mat{Y}\vect{v})
$$
subject to the constraints
$$Cov(\mat{X}\vect{w},\mat{X}\vect{w}_j) =0, ~ 1\leq j<h.$$
\end{remark}

The decomposition model for SIMPLS is identical to the decomposition of PLS2, the only difference being in how the weights $\vect{u}_h$ are calculated. In both models we have $\mat{\Xi}_{\bullet h} = \mat{X}\mat{W}_{\bullet h}$ (or $\mat{\Xi}_{\bullet h} = \mat{X}\tilde{\mat{W}}_{\bullet h}$), but the different constraints on the adjusted weights $\vect{w}_h$ (or $\tilde{\vect{w}}_h$) give different score vectors; $\|\vect{w}_h\|_2 = 1$ versus $\|\mathcal{P}_{\tilde{\mat{W}}_{\bullet h-1}^{\perp}}\tilde{\vect{w}}_h\|_2 = 1$. The predictions for both models are generated via $\widehat{\mat{Y}}_h = \mathcal{P}_{\mat{\Xi}_{\bullet h}}\mat{Y}$ 
so they will produce different predictions. \\


\begin{remark}
Another closely related (to SIMPLS) algorithm is the PLS simple iteration algorithm \cite{Zhu1995}. It has exactly the same objective function (and thus gives the same results) but differs in the way the matrices $\mathcal{P}_{\mat{C}_{\bullet h-1}}$ are computed. For the latter, the recursion formula $\mat{M}_h=\mat{M}_{h-1}-\mathcal{P}_{\mat{M}_{h-1}\mat{X}\transp\mat{X}\vect{w}_h}$ is used.
\end{remark}

\end{itemize}

\section{Penalized PLS}\label{Section3} 

All of the previous PLS methods can be written in terms of a single optimisation problem coupled with an appropriate deflation to ensure the appropriate orthogonal constraints. In this section we introduce the framework for penalised partial least squares in the unified PLS methodology. Several penalisations are then considered and presented in a unified algorithm that can preform all four PLS methods, and their regularised versions.

\subsection{Finding the PLS weights}
The $h$-th pair of penalised PLS weight vectors are given by the algorithm in Section C, where $P_{\lambda_1}$ and $P_{\lambda_2}$ are convex penalty functions with tuning parameters $\lambda_1$ and $\lambda_2$, and the matrix $\mat{M}_{h-1}$ is defined in the appropriate subsection of Section E. The resulting objective function solved at each iteration is convex in $\tilde{\vect{u}}$ (with fixed $\vect{v}$) and convex in $\tilde{\vect{v}}$ (for fixed $\vect{u}$). For a fixed unit norm $\vect{v}$, using the SVD connection, the optimisation is
\begin{equation}\label{pPLSu}
\begin{split}
\tilde{\vect{u}}_h = &\underset{\tilde{\vect{u}}}{\argmin}~\left\{\|\mat{M}_{h-1} - \tilde{\vect{u}}\vect{v}\transp\|_F^2 + P_{\lambda_1}(\tilde{\vect{u}})\right\},
\end{split}
\end{equation}
and we set $\vect{u}_h = \tilde{\vect{u}}_h/\|\tilde{\vect{u}_h}\|_2$ if $\|\tilde{\vect{u}_h}\|_2>0$ and $\vect{u}_h = \mathbf{0}_p$ otherwise. Similarly, for a fixed unit norm $\vect{u}$ we solve the optimisation problem
\begin{equation}\label{pPLSv}
\begin{split}
\tilde{\vect{v}}_h = &\underset{\tilde{\vect{v}}}{\argmin}~\left\{\|\mat{M}_{h-1}\transp - \tilde{\vect{v}}\vect{u}\transp\|_F^2 + P_{\lambda_2}(\tilde{\vect{v}})\right\},
\end{split}
\end{equation}
and set $\vect{v}_h = \tilde{\vect{v}}_h/\|\tilde{\vect{v}_h}\|_2$ if $\|\tilde{\vect{v}_h}\|_2>0$ and $\vect{v}_h = \mathbf{0}_q$ otherwise. For certain penalty functions, the convergence of this algorithm has been studied \cite{Witten2009}.

\subsection{Deflation and the PLS weights}

Computing the penalised versions of the four different PLS methods is achieved by alternating between two subtasks: solving (\ref{pPLSu}) and (\ref{pPLSv}) for the weights, and matrix deflation. Without the penalties $P_{\lambda_1}$ and $P_{\lambda_2}$, the matrix deflation enforces certain orthogonality constraints for each of the four standard PLS methods. However, with either penalty  $P_{\lambda_1}$ or $P_{\lambda_2}$, these deflations do not ensure any orthogonal constraints. Although, these constraints are lost, Witten et al. \cite{Witten2009}, state that it is not clear that orthogonality is desirable as it may be at odds with sparsity. That is, enforcing the additional orthogonality constraints may result in less sparse solutions. Similar to \cite{Witten2009,LeCao2008a} we use the standard deflation methods in our implementation of the penalised PLS methods. Alternative matrix deflations have been proposed for sparse PCA \cite{Mackey08}. However, these methods have not been extended in the general penalised PLS framework.

Another key observation is that for the NIPALS PLS regression, PLS-W2A, and CCA the scores were defined in terms of the deflated matrices $\xi_h = \mat{X}_{h-1}\vect{u}_h$ and $\omega_h = \mat{Y}_{h-1}\vect{v}_h$. Consequently the sparse estimates given by solving \eqref{pPLSu} and \eqref{pPLSv} perform variable selection of the deflated matrices. Thus the latent components formed using these methods have the interpretation given by Remark \ref{LatIntp}. In our implementation, we also calculate the adjusted weights $\vect{w}_h$ and $\vect{z}_h$ (or $\tilde{\vect{w}}_h$ and $\tilde{\vect{z}}_h$), where $\xi_h = \mat{X}\vect{w}_h$ and $\omega_h = \mat{Y}\vect{z}_h$. These weights allow for direct interpretation of the selected variables in the PLS model. Note that although $\vect{w}_h$ and $\vect{z}_h$ allow for direct interpretation of the selected variables, the sparsity is enforced on $\vect{u}_h$ and $\vect{v}_h$. So if $\vect{u}_h$ and $\vect{v}_h$ are sparse, this does not necessarily mean that the adjusted weights $\vect{w}_h$ and $\vect{z}_h$ will be sparse.

\begin{remark}\label{LatIntp}
It is important to understand how to interpret the results of such an analysis. The first latent variable $\vect{\xi}_1=\mat{X}\vect{u}_1$ is built as a sparse linear combination (with weights in $\vect{u}_1$) of the original variables. The next latent variable $\vect{\xi}_2=\mathcal{P}_{\vect{\xi}_1^\perp}\mat{X}\vect{u}_2$ is the part of the sparse linear combination (with weights in $\vect{u}_2$) of the original variables that has not been already explained by the first latent variable. And more generally, the $h$-th latent variable is built as a sparse linear combination of the original variables, from which we extract (by projection) the information not already brought by the previous latent variables.
\end{remark}

We note that an alternative SIMPLS formulation for the penalised PLS methods was proposed in a regression setting by \cite{Allen2013}. In the SIMPLS method the weights are directly interpreted in terms of the original variables, so $\vect{w}_h = \vect{u}_h$ and $\vect{z}_h = \vect{v}_h$. Although this method allows for direct penalisation of the weights, the orthogonality conditions still do not hold. We have incorporated this method and a similar variant for PLS-W2A in our package \texttt{bigsgPLS} to allow for direct penalisation of the weights.

\subsection{The penalised PLS methods}
Computationally, the PLS method is an efficient approach to sparse latent variable modelling. The main computational cost is in solving for the PLS weights as described in equations (\ref{pPLSu}) and (\ref{pPLSv}). The cost of solving for these weights is penalty specific but can is minimal in a number of useful applications. We detail a few examples where these equations have been solved analytically, and provide an algorithm that treats the penalised versions of the four PLS cases (i) -- (iv). 

\subsubsection{Sparse PLS} 

The (original) sparse PLS version sPLS \cite{LeCao2008a} (see also \cite{Chun2010}) considers the following penalty functions
\begin{equation} \label{l1pen}
P_{\lambda_1}(\tilde{\vect{u}}) = \sum_{i=1}^p2\lambda_1|\tilde{u}_i|\quad\textrm{ and }\quad P_{\lambda_2}(\tilde{\vect{v}}) = \sum_{j=1}^q 2\lambda_2|\tilde{v}_j|.
\end{equation}
These penalties induce the desired sparsity of the weight vectors $\vect{u}_h = \tilde{\vect{u}}_h / \|\tilde{\vect{u}}_h\|_2$ and $\vect{v}_h = \tilde{\vect{v}}_h / \|\tilde{\vect{v}}_h\|_2$, thanks to the well known properties of the $\ell_1$-norm or Lasso penalty \cite{Tibshirani1994}. The closed form solution for this problem is (see Appendix \ref{sPLSWeights} for proof):
\begin{equation}
\begin{split}
\tilde{\vect{u}} &= g^{\textrm{soft}}(\mat{M}\vect{v},{\lambda_1}),\\
\tilde{\vect{v}} &= g^{\textrm{soft}}(\mat{M}\transp\vect{u},{\lambda_2}).
\end{split}
\end{equation}
where $g^{\textrm{soft}}(\cdot,{\lambda_1})$ is the soft thresholding function, with the understanding that the function is applied componentwise. To unify these results with the ones to come, we introduce the sparsifyer functions $S_u$ and $S_v$ to denote analytical functions that provide the solution for the weights. The sparsifiers are functions of the data $\mat{M}$, the fixed weight $\vect{u}$ (or $\vect{v}$) and additional penalty specific parameters $\theta_u$ (or $\theta_v$). So for sparse PLS we have,
\begin{equation}
\begin{split}
\tilde{\vect{u}}_h &= S_u(\vect{v}; \mat{M}, \theta_u) = g^{\textrm{soft}}(\mat{M}\vect{v},{\lambda_1}),\\
\tilde{\vect{v}}_h &= S_v(\vect{u}; \mat{M}, \theta_v) = g^{\textrm{soft}}(\mat{M}\transp\vect{u},{\lambda_2}),
\end{split}
\end{equation}
where $\theta_u = \lambda_1$ and $\theta_2 = \lambda_2$.


\subsubsection{Group PLS}

There are many statistical problems in which the data has a natural grouping structure. In these problems, it is preferable to estimate all coefficients within a group to be zero or nonzero simultaneously. A leading example is in gene expression data, where genes within the same gene pathway have a similar biological function. Selecting a group amounts to selecting a pathway. Variables can be grouped for other reasons. For example, when we have categorical covariates in our data. This data is coded by their factor levels using dummy variables, and we desire selection or exclusion of this group of dummy variables.

Let us consider a situation where both matrices $\mat{X}$ and $\mat{Y}$ can be divided respectively into $K$ and $L$ sub-matrices (i.e., groups) $\mat{X}^{(k)}:n\times p_k$ and $\mat{Y}^{(l)}:n\times q_l$, where  $p_k$ (resp. $q_l$) is the number of covariates in group $k$ (resp. $l$). The aim is to select only a few groups of $\mat{X}$ which are related to a few groups of $\mat{Y}$. We define $\mat{M}^{(k,\cdot)} = \mat{X}^{(k)\transp}\mat{Y}$ and $\mat{M}^{(\cdot,l)} = \mat{Y}^{(l)\transp}\mat{X}$. 

Group selection is accomplished using the group lasso penalties \cite{Yuan2006} in the optimisation problems \eqref{pPLSu} and \eqref{pPLSv}:
\begin{equation}
\begin{split}
P_{\lambda_1}(\tilde{\vect{u}}) &= \lambda_1\sum_{k=1}^K\sqrt{p_k}\|\tilde{\vect{u}}^{(k)}\|_2^{};\\
P_{\lambda_2}(\tilde{\vect{v}}) &= \lambda_2\sum_{l=1}^L\sqrt{q_l}\|\tilde{\vect{v}}^{(l)}\|_2^{},
\end{split}
\end{equation}
where $\tilde{\vect{u}}^{(k)}$ and $\tilde{\vect{v}}^{(l)}$ are the sub vectors of the (unscaled) weights $\tilde{\vect{u}}$ and $\tilde{\vect{v}}$ corresponding to the variables in group $k$ of $\mat{X}$ and group $l$ of $\mat{Y}$ respectively. This penalty is a group generalisation of the Lasso penalty. Depending on the tuning parameter $\lambda_1 \geq 0$ (or $\lambda_2 \geq 0$), the entire weight subvector $\tilde{\vect{u}}^{(k)}$ (or $\tilde{\vect{v}}^{(l)}$) will be zero, or nonzero together.

The closed form solution for the group PLS method for the $k$-th subvector of $\tilde{\vect{u}}$ is given by
\begin{equation}\label{updateu}
S_u^{(k)}(\tilde{\vect{v}} ; \mat{M} , \vect{\theta}_u) = \left(1-\frac{\lambda_1}{2}\frac{\sqrt{p_k}}{\|\mat{M}^{(k,\cdot)}\tilde{\vect{v}}\|_2}\right)_+\mat{M}^{(k,\cdot)}\tilde{\vect{v}},
\end{equation}
so $\tilde{\vect{u}}^{(k)} = S_u^{(k)}(\tilde{\vect{v}} ; \mat{M} , \vect{\theta}_u)$. Similarly, the closed form solution for the $l$-th subvector of $\tilde{\vect{v}}$ is 
\begin{equation}\label{updatev}
S_v^{(l)}(\tilde{\vect{u}} ; \mat{M} , \vect{\theta}_v) = \left(1-\frac{\lambda_2}{2}\frac{\sqrt{q_l}}{\|\mat{M}^{(\cdot,l)}\tilde{\vect{u}}\|_2}\right)_+\mat{M}^{(\cdot,l)}\tilde{\vect{u}},
\end{equation}
so $\tilde{\vect{v}}^{(l)} = S_v^{(l)}(\tilde{\vect{u}} ; \mat{M} , \vect{\theta}_v)$. The sparsifyer functions are applied groupwise
\begin{equation*}
\begin{split}
\tilde{\vect{u}} &= S_u(\tilde{\vect{v}} ; \mat{M} , \vect{\theta}_u) = \left(S_u^{(1)}(\tilde{\vect{v}} ; \mat{M} , \vect{\theta}_u),\ldots,S_u^{(K)}(\tilde{\vect{v}} ; \mat{M} , \vect{\theta}_u)\right)\\
\tilde{\vect{v}} &= S_v(\tilde{\vect{u}} ; \mat{M} , \vect{\theta}_v) = \left(S_v^{(1)}(\tilde{\vect{u}} ; \mat{M} , \vect{\theta}_v),\ldots,S_v^{(L)}(\tilde{\vect{u}} ; \mat{M} , \vect{\theta}_v)\right),
\end{split}
\end{equation*}
with $\vect{\theta}_u = (p_1,\ldots,p_K,\lambda_1)$ and $\vect{\theta}_v = (q_1,\ldots,q_L,\lambda_2)$. A proof of these equations is given in \cite{Liquet2016}.

\subsubsection{Sparse Group PLS}

One potential drawback of gPLS is that it includes a group in the model only when all individual weights in that group are non-zero. However, sometimes we would like to combine both sparsity of groups and within each group. For example, if the predictor matrix contains genes, we might be interested in identifying particularly important genes in pathways of interest. The sparse group lasso \cite{Simon2013} achieves this within group sparsity. The sparse group selection in the PLS methodology is accomplished using the sparse group lasso penalty \cite{Simon2013} in the optimisation problem (\ref{pPLSu}) and (\ref{pPLSv}):
\begin{eqnarray*}
P_{\lambda_1}(\tilde{\vect{u}}) & = & (1-\alpha_1)\lambda_1\sum_{k=1}^K\sqrt{p_k}\|\tilde{\vect{u}}^{(k)}\|_2^{} +\alpha_1\lambda_1\|\tilde{\vect{u}}\|_1^{},\\
P_{\lambda_2}(\tilde{\vect{v}}) & = & (1-\alpha_2)\lambda_2\sum_{l=1}^L\sqrt{q_l}\|\tilde{\vect{v}}^{(l)}\|_2^{} +\alpha_2\lambda_2\|\tilde{\vect{v}}\|_1^{}.
\end{eqnarray*}
The sparse group penalty introduces tuning parameters $\alpha_2$ and $\alpha_2$ which provide a link between the group lasso penalty ($\alpha_1 = 0$, $\alpha_2 = 0$) and the lasso ($\alpha_1 = 1$, $\alpha_2 = 1$). Depending on the combination of $\alpha_1$ and $\lambda_1$ (or $\alpha_2$ and $\lambda_2$) the (unscaled) weight subvector $\tilde{\vect{u}}^{(k)}$ (or $\tilde{\vect{v}}^{(k)}$) will be eliminated entirely, or sparsely estimated. The adaptation of the sparse group penalty for the PLS method was first considered in \cite{Liquet2016}. The closed form solution of the sparse group PLS method for the $k$-th subvector of $\tilde{\vect{u}}^{(k)}$ is given by 

\begin{equation*}\label{condblocku}
S_u^{(k)}(\tilde{\vect{v}} ; \mat{M} , \vect{\theta}_u) = \begin{cases}
\vect{0} & \text{if }\frac{\|g_1\|_2}{(1-\alpha_1)\sqrt{p_k}}\leq\lambda_1\\
\frac{g_1}{2} -
\frac{\lambda_1(1-\alpha_1)\sqrt{p_k}g_1}{2\|g_1\|} & \text{otherwise}
\end{cases}
\end{equation*}
where $g_1 = g^{\textrm{soft}}\left(\mat{M}^{(k,\cdot)}\tilde{\vect{v}},\lambda_1\alpha_1/2\right)$. Similarly, the $l$-th subvector of $\tilde{\vect{v}}$ is given by
\begin{equation*}\label{condblockv}
S_v^{(l)}(\tilde{\vect{v}} ; \mat{M} , \vect{\theta}_v) = \begin{cases}
\vect{0} & \text{if }\frac{\|g_2\|_2}{(1-\alpha_2)\sqrt{q_l}}\leq\lambda_2\\
\frac{g_2}{2} -
\frac{\lambda_2(1-\alpha_2)\sqrt{q_l}g_2}{2\|g_2\|} & \text{otherwise}
\end{cases}
\end{equation*}
where $g_2 = g^{\textrm{soft}}\left(\mat{M}^{(\cdot,l)}\tilde{\vect{u}},\lambda_2\alpha_2/2\right)$. The sparsifyer functions for these penalties are:
\begin{equation*}
\begin{split}
\tilde{\vect{u}} = S_u(\tilde{\vect{v}} ; \mat{M} , \vect{\theta}_u) = \left(S_u^{(1)}(\tilde{\vect{v}} ; \mat{M} , \vect{\theta}_u),\ldots,S_u^{(K)}(\tilde{\vect{v}} ; \mat{M} , \vect{\theta}_u)\right)\\
\tilde{\vect{v}} = S_v(\tilde{\vect{u}} ; \mat{M} , \vect{\theta}_v) = \left(S_v^{(1)}(\tilde{\vect{u}} ; \mat{M} , \vect{\theta}_v),\ldots,S_v^{(L)}(\tilde{\vect{u}} ; \mat{M} , \vect{\theta}_v)\right)
\end{split}
\end{equation*}
with $\vect{\theta}_u = (p_1,\ldots,p_K,\lambda_1,\alpha_1)$ and $\vect{\theta}_v = (q_1,\ldots,q_L,\lambda_2,\alpha_2)$.\\

\subsubsection{Other penalties} The penalties discussed so far have enforced general sparsity or sparsity with respect to a known grouping structure in the data. Extensions to the group structured sparsity in partial least squares setting have also been considered; in terms of overlapping groups \cite{Chen2012}, or additional grouping restrictions \cite{sutton2017}. The penalisations considered so far all have all resulted in closed form solutions for the updates of $\vect{u}$ and $\vect{v}$. We note here that this is not always the case. The fused Lasso penalty \cite{Tibshirani2005} is defined by:
\begin{eqnarray*}
P_{\lambda_1}(\tilde{\vect{u}}) & = & \lambda_1\alpha_1\sum_{i=2}^p|\tilde{u}_i - \tilde{u}_{i-1}| +(1-\alpha_1)\lambda_1\|\tilde{\vect{u}}\|_1^{},\\
P_{\lambda_2}(\tilde{\vect{v}}) & = & \lambda_1\alpha_1\sum_{i=2}^q|\tilde{v}_i - \tilde{v}_{i-1}| +(1-\alpha_1)\lambda_1\|\tilde{\vect{v}}\|_1^{}.
\end{eqnarray*}
The first term in this penalty causes neighbouring coefficients to shrink together and will cause some to be identical, and the second causes regular Lasso shrinkage of the parameters for variable selection. Unlike the previous methods, a closed form solution for the fused Lasso cannot be directly achieved. This is because the penalty is not a separable function of the coordinantes. Because there is no closed form solution for the fused Lasso, we cannot write a sparsifyer function so we have not considered this method. We note that methods exist that are able to solve the fused Lasso problem, either by reparameterisation, dynamic programming or path based algorithms. In particular, \cite{Witten2009} have considered solving problems of the form (\ref{pPLSu}) and \eqref{pPLSv} with the fused Lasso penalty. In their paper, they propose a sparse and fused penalised CCA, however in their derivation they assume $\mat{X}\transp\mat{X} = \mat{I}$ and $\mat{Y}\transp\mat{Y}= \mat{I}$. In our framework, this method would be sparse and fused penalised PLS-SVD.

\section{The unified algorithm}\label{Section4}


Algorithm~1 permits to compute in a unified way, all four PLS versions (i)--(iv), with a possibility to add sparsity. Adjusted weights can also be computed and, if the number of requested components $H$ is greater than 1, a deflation step is executed. Note that, if $\mat{Y}$ is taken equal to $\mat{X}$, this algorithm performs Principal Component Analysis (PCA), as well as sparse PCA versions. If this is the case, the optimized criteria are simply restated in terms of variance instead of covariance.





\begin{center}
ALGORITHM~1 HERE
\end{center}

\begin{algorithm}
\begin{algorithmic}[1]
\Require $\lambda_x$, $\lambda_y$, $H$, $\mat{X}_0 = \mat{X}$, $\mat{Y}_0 = \mat{Y}$, $\vect{\theta}_u$, $\vect{\theta}_v$
\State $\mat{M}_0 \leftarrow \mat{X}_0\transp\mat{Y}_0$ \Comment{Initialisation}
\State $\mat{P} \leftarrow \mat{I}$ and $\mat{Q} \leftarrow \mat{I}$
\State $\vect{u}_{0} \leftarrow  \vect{c}_{0}   \leftarrow \vect{0}_p$, $\vect{v}_{0} \leftarrow  \vect{0}_q$ and $\vect{\xi}_0 \leftarrow \vect{\omega}_0 \leftarrow \vect{1}_{n}$
\State \textbf{If} Case (iii) \textbf{then} 
\State ~~~~~~ $\mat{A} \leftarrow (\mat{X}_0\transp\mat{X}_0+\lambda_x\mat{I})^{-1/2}$
\State ~~~~~~ $\mat{B}\leftarrow (\mat{Y}_0\transp\mat{Y}_0+\lambda_y\mat{I})^{-1/2}$
\State ~~~~~~ $\mat{M}_0 \leftarrow \mat{A}\mat{M}_0\mat{B}$
\State \textbf{end if}
\For {$h = 1,\ldots, H$}
\State Apply the SVD to $\mat{M}_{h-1}$ and extract the first 
\State triplet $(\delta_1,\vect{u}_1,\vect{v}_1)$ of singular value and vectors.
\State Set $\vect{u}_{\textrm{h}} \leftarrow \vect{u}_1$ and $\vect{v}_{\textrm{h}} \leftarrow \vect{v}_1$
\While {convergence$^{(*)}$ of $\vect{u}_{h}$}  \Comment{Sparsity step}
\State $\tilde{\vect{u}}_{h} \leftarrow S_u(\vect{v}_{h} ; \mat{M}_{h-1} , \vect{\theta}_u)$
\State $\vect{u}_{h} \leftarrow \tilde{\vect{u}}_{h} / \|\tilde{\vect{u}}_{h}\|_2^{}$
\State $\tilde{\vect{v}}_{h} \leftarrow S_v(\vect{u}_{h} ; \mat{M}_{h-1} , \vect{\theta}_v)$
\State $\vect{v}_{h} \leftarrow \tilde{\vect{v}}_{h} / \|\tilde{\vect{v}}_{h}\|_2^{}$
\EndWhile \Comment{End of sparsity step}
\State $\vect{\xi}_h \leftarrow \mat{X}_{h-1}\vect{u}_{h}$ \Comment{X-score}
\State $\vect{\omega}_h \leftarrow \mat{Y}_{h-1}\vect{v}_{h}$ \Comment{Y-score}

\State \textbf{If} Case (i) \textbf{then} \Comment{Adjusted weights step}
\State ~~~~~~ $\vect{w}_{h} \leftarrow \vect{u}_{h}$ and $\vect{z}_{h} \leftarrow \vect{v}_{h}$ 
\State \textbf{end if}
\State \textbf{If} Case (ii) \textbf{then} 
\State ~~~~~~ $\mat{P} \leftarrow \mat{P}(\mat{I} - \vect{u}_{h-1}\vect{\xi}_{h-1}\transp\mat{X}/\|\vect{\xi}_{h-1}\|^2)$
\State ~~~~~~ $\mat{Q} \leftarrow \mat{Q}(\mat{I} - \vect{v}_{h-1}\vect{\omega}_{h-1}\transp\mat{X}/\|\vect{\omega}_{h-1}\|^2)$
\State ~~~~~~ $\vect{w}_h \leftarrow \mat{P}\vect{u}_h$ and $\vect{z}_h \leftarrow \mat{Q}\vect{v}_h$
\State \textbf{end if}
\State \textbf{If} Case (iii) \textbf{then} $\vect{w}_{h} \leftarrow \mat{A}\vect{u}_{h}$ and $\vect{z}_{h} \leftarrow \mat{B}\vect{v}_{h}$ 
\State \textbf{If} Case (iv) \textbf{then} 
\State ~~~~~~ $\mat{P} \leftarrow \mat{P}(\mat{I} - \vect{u}_{h-1}\vect{c}_{h-1}\transp)$
\State ~~~~~~ $\vect{w}_h \leftarrow \mat{P}\vect{u}_h$
\State ~~~~~~ $\vect{z}_h \leftarrow \vect{v}_h$ 
\State \textbf{end if} \Comment{End of adjusted weights step}
\State \textbf{If} Case (i) or (iii) \textbf{then}  \Comment{Deflation step}
\State ~~~~~~ $\vect{c}_h\transp \leftarrow \vect{u}_h\transp$ and $\vect{e}_h\transp \leftarrow \vect{v}_h\transp$
\State \textbf{end if}
\State \textbf{If} Case (ii) or (iv) \textbf{then} $\vect{c}_h\transp \leftarrow \vect{\xi}_h\transp\mat{X}_{h-1}/\|\vect{\xi}_h\|_2^2$ 
\State \textbf{If} Case (ii) \textbf{then} $\vect{e}_h\transp \leftarrow \vect{\omega}_h\transp\mat{Y}_{h-1}/\|\vect{\omega}_h\|_2^2$ 
\State \textbf{If} Case (iv) \textbf{then} $\vect{d}_{h}\transp \leftarrow \vect{\xi}_h\transp\mat{Y}_{h-1}/\|\vect{\xi}_h\|_2^2$ 
\State  $\mat{X}_{h} \leftarrow \mat{X}_{h-1}-\vect{\xi}_h\vect{c}\transp_h$ 
\State \textbf{If} Case (iv) \textbf{then} 
\State ~~~~~~ $\mat{Y}_{h} \leftarrow \mat{Y}_{h-1} - \vect{\xi}_h\vect{d}\transp_{h}$ \Comment{PLS2}
\State \textbf{Else}
\State ~~~~~~ $\mat{Y}_{h} \leftarrow \mat{Y}_{h-1} - \vect{\omega}_h\vect{e}\transp_h$ 
\State \textbf{End If}
\State $\mat{M}_h \leftarrow \mat{X}_h\transp\mat{Y}_h$  \Comment{End of deflation step} \\
\textbf{Store} $\vect{\xi}_h$, $\vect{\omega}_h$, $\vect{u}_h$, $\vect{v}_h$, $\vect{w}_h$, $\vect{z}_h$
\EndFor
\end{algorithmic} 
\caption{Sparse PLS algorithm for the four cases (i)--(iv)}
\label{alg-opt}
$(*)$ Convergence of a vector $\vect{t}$ is tested on the change in $\vect{t}$, i.e., $\|\vect{t}_{\text{old}}-\vect{t}_{\text{new}}\|/\|\vect{t}_{\text{new}}\|<\epsilon$, where $\epsilon$ is ``small'', e.g., $10^{-8}$.
\end{algorithm}

\begin{remark}
On line~10, we impose that $\|\vect{u}_1\|_2^{} = \|\vect{v}_1\|_2^{} = 1$ and $u_{1,i}>0$ where $i=\argmax_{1\leq j\leq p}|u_{1,j}|$ to ensure uniqueness of the results. \\
Note that $\vect{w}_h$ and $\vect{z}_h$ of lines~27, 29 and 32 correspond to $\tilde{\vect{w}}_h$ and $\tilde{\vect{z}}_h$ in the text.
\end{remark}
At this point, it is worthwhile noting that that when $p$ and $q$ are small compared to $n$, one can slightly modify Algorithm~1 by using the recursive equations that express $\mat{M}_h$ in terms of $\mat{M}_{h-1}$, instead of using the recursions on $\mat{X}_h$ and $\mat{Y}_h$. The former are provided in subsection~\ref{fourstandardPLS}. This should increase speed of execution of the algorithm. 

Moreover, one can use various approaches to deal with the cases when $n$, $p$ or $q$ are too large in our algorithm, making some objects not fittable into the computer's memory. These can be divided into \textit{chunk approaches} and \textit{streaming (or incremental) approaches}, which are presented in the next subsections. Of course, any combinations of these approaches can be used if necessary. Some of these approaches might even increase the computation speed, even in a context where all objects would fit into memory.

\subsection{Matrix multiplication using chunks}

To scale Algorithm~1 to big data (i.e., very large $n\gg p\  \textrm{and}\ q$), we can use a simple idea to multiply two very large matrices that are too big to fit into the computer's memory. 

Let us divide the total number $n$ of rows of $\mat{X}$ (resp. of $\mat{Y}$) into blocks $\mat{X}_{(g)}$ (resp. $\mat{Y}_{(g)}$), $g=1,\ldots,G$, of (approximatively) the same size. We have
$$
\mat{X}\transp\mat{Y} = \sum_{g=1}^G\mat{X}_{(g)}\transp\mat{Y}_{(g)}.
$$
The number of blocks $G$ has to be chosen so that each product $\mat{X}_{(g)}\transp\mat{Y}_{(g)}:p\times q$ can be done within the available RAM. Note that all these products can be performed in parallel if the required computing equipment is available.



\subsection{SVD when $p$ or $q$ is very large}

The main step of our algorithm is the computation  of the  first triplet $(\delta_1,\vect{u}_1,\vect{v}_1)$ in the SVD of the $(p\times q)$ matrices $\mat{M}_{h-1}$. The \texttt{irlba} \cite{irlba} R package can be used to compute quite easily this triplet for values of $p$ and $q$ as big as $50,000$. This package is based on an augmented implicitly restarted Lanczos bidiagonalization method \cite{Baglama05b}.



When $p$ (or $q$) is much larger, another approach is necessary to compute the SVD of $\mat{M}_{h-1}$; see e.g., \cite{LIA16}. Suppose that $p$ is large but not $q$, which is common in several applications. We thus suppose that $p\gg q$. The Algorithm~1 in \cite{LIA16} is now presented to highlight the elements needed in our algorithm. We can partition a large matrix $\mat{M}:p\times q$ by rows into a small number $s$ of submatrices (or chunks):
$$
\mat{M} = \left(
\begin{array}{c}
\mat{M}_1\\
\mat{M}_2\\
\vdots\\
\mat{M}_s
\end{array}
\right).
$$
Let $\mat{M}_i = \mat{U}_i\mat{D}_i\mat{V}_i\transp$ denote the SVD of $\mat{M}_i:g\times q$ such that $gs=p$ (w.l.o.g.). We can take $g$ much larger than $q$ as long as it is still possible to compute the SVD of these submatrices. Define 
$$
\tilde{\mat{U}} = \left(
\begin{array}{cccc}
\mat{U}_1 & & & \\
          & \mat{U}_2 & & \\
          &           & \ddots & \\
          &           &        & \mat{U}_s
\end{array}
\right):p\times p
$$
where $\mat{U}_i:g\times g$ and define
$$
\mat{H} = \left(
\begin{array}{c}
\mat{D}_1\mat{V}_1\transp\\
\mat{D}_2\mat{V}_2\transp\\
\vdots\\
\mat{D}_s\mat{V}_s\transp\\
\end{array}
\right):p\times q,
$$
where $\mat{D}_i:g\times q$ and $\mat{V}_i:q\times q$, so that $\mat{M} = \tilde{\mat{U}}\mat{H}$. 
Let $\mat{H} = \mat{U}_H\mat{D}_H\mat{V}_H\transp$ be the SVD of $\mat{H}$. Note that this matrix is as large as $\mat{M}$ so one may wonder what has been gained with this approach. But $\mat{D}_i$ being a $g\times q$ diagonal rectangular matrix, $\mat{D}_i\mat{V}_i\transp$ has $g-q$ zero row-vectors in its bottom. Consequently, the matrix $\mat{H}$ contains only $sq$ non-zero row vectors. Now let $\tilde{\mat{H}} = \mat{R}\mat{H}:p\times q$ be a rearrangement in rows for $\mat{H}$ such that its first $sq$ row vectors are non-zero and $p-sq$ row vectors are in its bottom. We now have to compute $\mat{U}^*\mat{D}^*\mat{V}^{*\transp}$, the SVD of a (much smaller) $sq\times q$ matrix\footnote{The transpose sign on $\mat{R}$ is missing in \cite{LIA16}.}:
\begin{eqnarray*}
\mat{H} & = & \mat{R}\transp\tilde{\mat{H}} \\
& = & \mat{R}\transp\left[\begin{array}{c}\mat{U}^*\mat{D}^*\mat{V}^{*\transp}\\
\boldsymbol{0}\end{array}\right] \\
& = & \underbrace{\mat{R}\transp\left[\begin{array}{cc}\mat{U}^* & \boldsymbol{0}\\\boldsymbol{0} & \mat{I}_{p-sq}\end{array}\right]}_{\mat{U}_H}\underbrace{\left[\begin{array}{c}\mat{D}^*\\\boldsymbol{0}\end{array}\right]}_{\mat{D}_H}\underbrace{\mat{V}^{*\transp}}_{\mat{V}_H\transp}.
\end{eqnarray*}
We obtain
$$
\mat{M} = (\tilde{\mat{U}}\mat{U}_H)\mat{D}_H\mat{V}_H\transp
$$
which forms a SVD of $\mat{M}$.

Now, let $\boldsymbol{1}$ be a vector containing only 0s but a 1 in the first position. For our PLS algorithm, we only need to compute the first triplet in the SVD of $\mat{M}$, namely $\delta_1 = \boldsymbol{1}_p\transp\mat{D}_H\boldsymbol{1}_q = \mat{D}^*_{1,1}$, $\vect{v}_1  = \mat{V}_{\bullet 1} = \mat{V}_H\boldsymbol{1}_q = \mat{V}^*\boldsymbol{1}_q$ and the first column of $(\tilde{\mat{U}}\mat{U}_H)$:
\begin{eqnarray*}
\vect{u}_1=(\tilde{\mat{U}}\mat{U}_H)\boldsymbol{1}_p & = & \tilde{\mat{U}}\mat{R}\transp\left[\begin{array}{cc}\mat{U}^* & \boldsymbol{0}\\\boldsymbol{0} & \mat{I}_{p-sq}\end{array}\right]\boldsymbol{1}_p\\
& = & \tilde{\mat{U}}\mat{R}\transp\left[\begin{array}{c}\mat{U}^*\boldsymbol{1}_{sq}\\\boldsymbol{0}\end{array}\right]\\
& = & \left[
\begin{array}{c}
\mat{U}_{1,\bullet q}(\mat{U}^*_{\bullet 1})_{1,\ldots,q} \\
\mat{U}_{2,\bullet q}(\mat{U}^*_{\bullet 1})_{q+1,\ldots,2q} \\
\vdots\\
\mat{U}_{s,\bullet q}(\mat{U}^*_{\bullet 1})_{(s-1)q+1,\ldots,sq} \\
\end{array}
\right].
\end{eqnarray*}
It is seen above that only the $q$ first triplets of the SVDs of the $\mat{M}_i$s are required. So, overall we ``only'' have to compute $s$ truncated ($q\times q$) SVDs (of the $\mat{M}_i$s) and one truncated ($1\times 1$) SVD (of the $sq$ first lines of $\tilde{\mat{H}}$, which are easily obtained from these truncated SVDs).


Moreover, we can even compute $\vect{u}_1$ from $\vect{v}_1$ using the simple formula $\vect{u}_1 = \mat{M}\vect{v}_1 / \|\mat{M}\vect{v}_1\|$ (using a chunk approach).\\

When $q$ is larger than $p$, we just partition $\mat{M}$ in columns instead of rows. 
When both $p$ and $q$ are large, one can adapt Algorithm~2 in \cite{LIA16} which generalizes the above. (They even propose a third algorithm for the case of online (streaming) SVDs.)

Note that these algorithms based on the split-and-merge strategy possess an embarrassingly parallel structure and thus can be efficiently implemented on a distributed or multicore machine.







\subsection{Incremental SVD when $n$ is large}

We want to compute the truncated SVD 
of the matrix $\mat{M}_h = \mat{X}_h\transp\mat{Y}_h$ when $n$ is very large (and the $X$- and $Y$-matrices are split in blocks, or chunks, of size $n/G$ for some given $G$). One can use the divide and conquer approach presented in subsection~A to compute first the matrix $\mat{M}_h = \mat{X}_h\transp\mat{Y}_h$ and then evaluate the SVD of this matrix. We present here an alternative approach \cite{Cardot2015} by considering an incremental version of the SVD.

Let $\mat{X}\transp = [\vect{x}_1\transp,\ldots,\vect{x}_n\transp]$ and $\mat{Y}\transp = [\vect{y}_1\transp,\ldots,\vect{y}_n\transp]$ be non-centered data matrices. We note
$$
\mat{M}_n = \dot{\mat{X}}_n\transp\dot{\mat{Y}}_n = \sum_{i=1}^n(\vect{x}_i - \vect{\mu}_{X,n})(\vect{y}_i - \vect{\mu}_{Y,n})\transp
$$
with the centered data matrices
$$
\dot{\mat{X}}_n = \mat{X}_n - \vect{1}_n\vect{\mu}_{X,n}\transp,\quad \dot{\mat{Y}}_n = \mat{Y}_n - \vect{1}_n\vect{\mu}_{Y,n}\transp
$$
where
$$
\vect{\mu}_{X,n} = n^{-1}\mat{X}\transp\vect{1}_n = n^{-1}\sum_{i=1}^n\vect{x}_i\transp$$
and
$$
\vect{\mu}_{Y,n} = n^{-1}\mat{Y}\transp\vect{1}_n = n^{-1}\sum_{i=1}^n\vect{y}_i\transp.
$$
We have the streaming updating formulas
$$
\vect{\mu}_{X,n+1} = \frac{n}{n+1}\vect{\mu}_{X,n} + \frac{1}{n+1}\vect{x}_{n+1},
$$
$$
\vect{\mu}_{Y,n+1} = \frac{n}{n+1}\vect{\mu}_{Y,n} + \frac{1}{n+1}\vect{y}_{n+1},
$$
and
{\small\begin{eqnarray}\label{updateMn}
\mat{M}_{n+1} & = & \mat{M}_n + \frac{n}{(n+1)}(\vect{x}_{n+1} - \vect{\mu}_{X,n})(\vect{y}_{n+1} - \vect{\mu}_{Y,n})\transp.~~~~~~~
\end{eqnarray}}
Now, let the $H$-rank truncated SVD of $\mat{M}_n$ be $\mat{M}_n^{(H)} = \mat{U}_{n,\bullet H}\mat{\Delta}_{n,H}\mat{V}_{n,\bullet H}\transp$. Let $\tilde{\vect{x}}_{n+1} = \vect{x}_{n+1} - \vect{\mu}_{X,n}$ and $\tilde{\vect{y}}_{n+1} = \vect{y}_{n+1} - \vect{\mu}_{Y,n}$. Since $\mat{U}_{n,\bullet H}\transp\mat{U}_{n,\bullet H}=\mat{I}$, we have
\begin{eqnarray*}
\tilde{\vect{x}}_{n+1} & = & \mathcal{P}_{\mat{U}_{n,\bullet H}}\tilde{\vect{x}}_{n+1} + \mathcal{P}_{\mat{U}_{n,\bullet H}^\perp}\tilde{\vect{x}}_{n+1} \\
& = & \mat{U}_{n,\bullet H}\mat{U}_{n,\bullet H}\transp\tilde{\vect{x}}_{n+1} + \mathcal{P}_{\mat{U}_{n,\bullet H}^\perp}\tilde{\vect{x}}_{n+1} \\
& = & \mat{U}_{n,\bullet H}\vect{c}_{n+1} +  \tilde{\vect{x}}_{n+1}^\perp
\end{eqnarray*}
with $\vect{c}_{n+1} = \mat{U}_{n,\bullet H}\transp\tilde{\vect{x}}_{n+1}$ and $\tilde{\vect{x}}_{n+1}^\perp = \mathcal{P}_{\mat{U}_{n,\bullet H}^\perp}\tilde{\vect{x}}_{n+1}$. Similarly, 
$$
\tilde{\vect{y}}_{n+1} = \mat{V}_{n,\bullet H}\vect{d}_{n+1} + \tilde{\vect{y}}_{n+1}^\perp
$$
with $\vect{d}_{n+1} = \mat{V}_{n,\bullet H}\transp\tilde{\vect{y}}_{n+1}$ and $\tilde{\vect{y}}_{n+1}^\perp = \mathcal{P}_{\mat{V}_{n,\bullet H}^\perp}\tilde{\vect{y}}_{n+1}$. Now, in view of \eqref{updateMn}, we have the approximation 
$$
\mat{M}_{n+1}^{(H)} \approx \mat{M}_n^{(H)} + \frac{n}{n+1}\tilde{\vect{x}}_{n+1}\tilde{\vect{y}}_{n+1}\transp.
$$
\begin{remark}
Note that this approximation is in fact exact when $H=\textrm{rank}(\mat{M}_n)$. So if we want to use this approach in our algorithm, we would have to compute all the singular elements and not only the first triplet. This being said, if for example $q$ is not too large (e.g., $q=1$) this is not a problem anymore. Moreover, we see from Appendix~\ref{C1Proof} that $\vect{u}_1 = \mat{X}\transp\mat{Y}\vect{v}_1 / \|\mat{X}\transp\mat{Y}\vect{v}_1\|$ and $\vect{v}_1 = \mat{Y}\transp\mat{X}\vect{u}_1 / \|\mat{Y}\transp\mat{X}\vect{u}_1\|$. Note also that $\vect{u}_1$ is the first eigenvector of the $p\times p$ matrix $(\mat{Y}\transp\mat{X})\transp\mat{Y}\transp\mat{X}$ whereas $\vect{v}_1$ is the first eigenvector of the $q\times q$ matrix $(\mat{X}\transp\mat{Y})\transp\mat{X}\transp\mat{Y}$. So we only need to compute either $\vect{u}_1$ (if $p<q$) or $\vect{v}_1$ (if $q\leq p$), from which we obtain the other one.
\end{remark}
At this point, one can write
$$
\mat{M}_{n+1}^{(H)} = \left[\mat{U}_{n,\bullet H}, \frac{\tilde{\vect{x}}_{n+1}^\perp}{\|\tilde{\vect{x}}_{n+1}^\perp\|_2}\right]\mat{Q}_{n+1}\left[\mat{V}_{n,\bullet H}, \frac{\tilde{\vect{y}}_{n+1}^\perp}{\|\tilde{\vect{y}}_{n+1}^\perp\|_2}\right]\transp
$$
with
$$
\mat{Q}_{n+1} = \frac{n}{n+1}\left(
\begin{array}{cc}
\frac{n+1}{n}\mat{\Delta}_n + \vect{c}_{n+1}\vect{d}_{n+1}\transp & \|\tilde{\vect{y}}_{n+1}^\perp\|_2\vect{c}_{n+1} \\
\|\tilde{\vect{x}}_{n+1}^\perp\|_2\vect{d}_{n+1}\transp & \|\tilde{\vect{x}}_{n+1}^\perp\|_2\|\tilde{\vect{y}}_{n+1}^\perp\|_2
\end{array}
\right).
$$
It then suffices to perform the SVD of the matrix $\mat{Q}_{n+1}$ of dimension $(H+1)\times(H+1)$. Writing $\mat{Q}_{n+1} = \mat{A}_{n+1}\mat{S}_{n+1}\mat{B}_{n+1}\transp$, we have 
$$
\mat{M}_{n+1}^{(H)} = \mat{U}_{n+1}\mat{\Delta}_{n+1}\mat{V}_{n+1}\transp
$$
with
$\mat{\Delta}_{n+1} = \mat{S}_{n+1}$,
$$
\mat{U}_{n+1} = \left[\mat{U}_n,\frac{\tilde{\vect{x}}_{n+1}^\perp}{\|\tilde{\vect{x}}_{n+1}^\perp\|_2} \right]\mat{A}_{n+1}
$$
and
$$
\mat{V}_{n+1} = \left[\mat{V}_n,\frac{\tilde{\vect{y}}_{n+1}^\perp}{\|\tilde{\vect{y}}_{n+1}^\perp\|_2} \right]\mat{B}_{n+1}.
$$
To keep the approximation $\mat{M}_{n+1}^{(H)}$ of $\mat{M}_{n+1}$ at rank $H$, the row and column of $\mat{\Delta}_{n+1}$ containing the smallest singular value are deleted and the associated singular vectors are deleted from $\mat{U}_{n+1}$ and $\mat{V}_{n+1}$.\\

This incremental way to compute the SVD provides a promising alternative for handling very large sample size (specially when $q$ is not too large). Moreover the incremental SVD is well designed in a data stream context.

\section{Numerical Experiments}\label{Section5}

 In this section we use the R software to carry out a short simulation study 
in order to illustrate the numerical behaviour of the new proposed approach. The experiments have been conducted using a laptop with a 2.53~GHz
processor and 8~GB of memory. The parallel strategy utilizes four processor cores.
 
We present two  simulations to illustrate the good performance of the proposed approaches and the scalability to large sample sizes of our algorithm.
The first simulation considers the PLS-R model (case (iv)) on group structure data while the second simulation presents an extension of PLS approaches  to discriminant analysis purpose.

\subsection{Group PLS model}
We generate data with a group structure:  20 groups of 20 variables for  $\boldsymbol{X}$ ($p=400$) and 25 groups of 20 variables for  $\boldsymbol{Y}$ ($q=500$).
To highlight the scalability of our algorithm, we generate two big matrices from the following  models linked by $H=2$ latent variables:
\begin{equation}\label{simu1}
\mat{X} =  \mat{\Xi}_{H}\mat{C}_{H}\transp + \mat{F}_{H}^{X},\qquad\mat{Y} =  \mat{\Xi}_{H}\mat{D}_{H}\transp + \mat{F}_{H}^{Y},
\end{equation}
where the matrix $\mat{\Xi}_H =(\vect{\xi}_j)$ contains 2 latent variables $\vect{\xi}_1$ and $\vect{\xi}_2$. The entries in these vectors have all been independently generated from a standard normal distribution. The rows of the residual
matrix $\mat{F}_{H}^{X}$ (respectively, $\mat{F}_{H}^{Y}$) have been generated from a multivariate normal distribution with zero mean $\vect{\mu}_X$ (resp. $\vect{\mu}_Y$) and covariance matrix $\mat{\Sigma}_X=1.5^2\mat{I}_p$ (resp. $\mat{\Sigma}_Y=1.5^2\mat{I}_q$).

Among the 20 groups of $\mat{X}$, only 4 groups each containing 15 true variables and 5 noise variables are associated to the response variables of $\mat{Y}$. We set the
$p$-vector $\vect{c}_1$ (first column of the  $\mat{C}_{H}$ matrix) to have 15 1s, 30 -1s and 15 1.5s, the other entries being all set to 0. All 15 non-zero coefficients are assigned randomly into one group along with the remaining 5 zero coefficients corresponding
to noise variables. The vector $\vect{c}_2$ is
chosen in the same way as $\vect{c}_1$. The two columns
of $\mat{D}_{H}$ are $q$-vectors containing 15 -1s, 15 -1.5s and 30 1s and
the rest are 0s such that the matrix $\mat{Y}$ have a similar group structure for 4 groups containing the signal. Finally, the sample size is set to $n=560,000$ observations which corresponds  to storage requirements of approximately 5~GB for each matrix, thus with a total exceeding the 8~GB of memory available on our laptop.

\begin{figure}[h] \centering
\includegraphics[width=8cm,height=8cm]{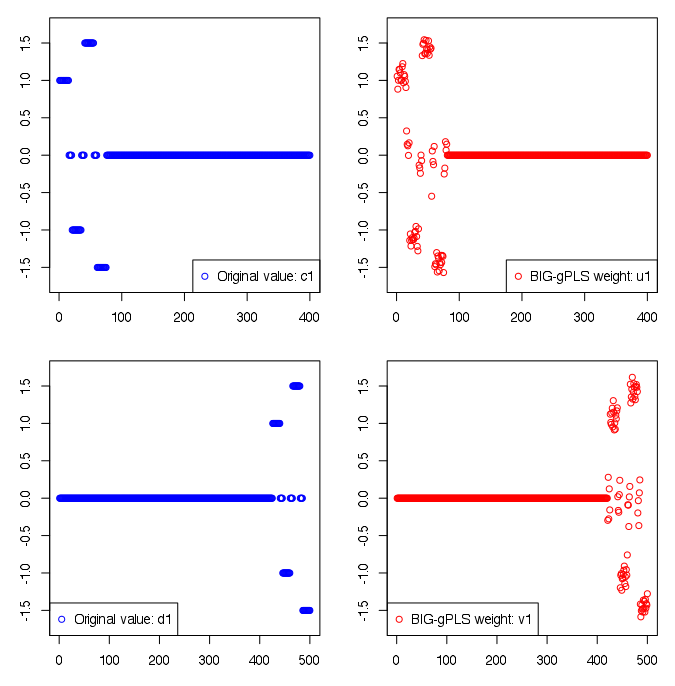}
\includegraphics[width=8cm,height=8cm]{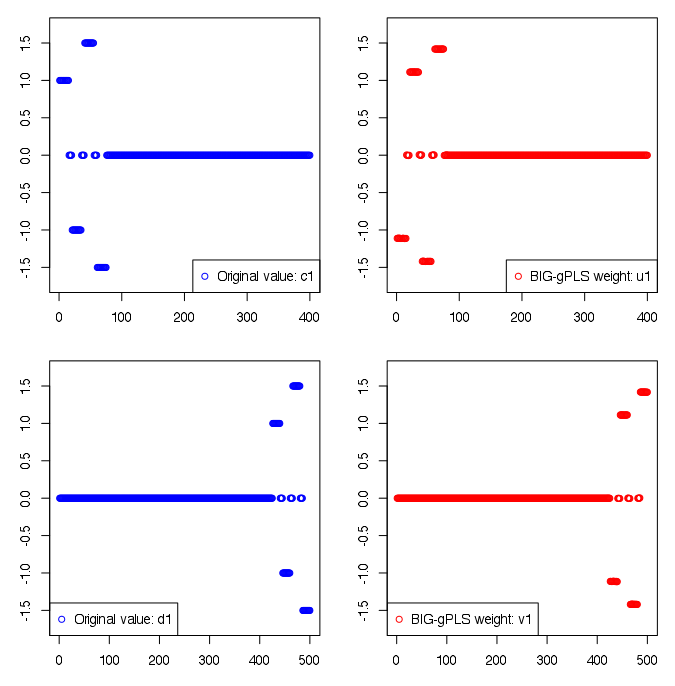}
\caption{Comparison of the signal recovered (weights $\vect{u}_1$ and $\vect{v}_1$) by
the first component ($H = 2$) of the gPLS. For the top four plots, $n=100$, and for the four bottom plots $n=560,000$. Left column: the true values of $c_1$ and $d_1$ for small and large sample sizes. Right column: the estimated values of $c_1$ and $d_1$ for small and large sample sizes. Note that the values of $\vect{u}_1$ (resp. $\vect{v}_1$) have
been rescaled so that its norm equals that of the original $\vect{c}_1$ (resp. $\vect{d}_1$)} \label{plotsimu1}
\end{figure}

The top four plots of Figure~\ref{plotsimu1} show the results of the group PLS estimated with only $n=100$ observations. For such a sample size, the usual group PLS can be used without any computational time or memory issues. In this case, group PLS manages to select the relevant groups and performs well to estimate
the weight vectors $\vect{u}_1$ and $\vect{v}_1$ related to the first component and the
weight vectors $\vect{u}_2$ and $\vect{v}_2$ related to the second component. 

The bottom four plots of Figure~\ref{plotsimu1} show the results of the group PLS estimated on the full data set which can be only analyzed by using the extended version of our algorithm for big data. In this run, we use $G=100$ chunks for enabling matrix multiplication. The execution time was around 15~minutes for two components ($H=2$) and took less than 2 minutes for getting the first component. We can observe that the signal has been perfectly identified and estimated, which is expected for such a huge amount of information.

Note that for validation purposes, the extended version of our algorithm for big data have been ran and gave exactly the same results than the usual algorithm on the small data set ($n=100$). 





\subsection{Case of Regularised  PLS-DA}

\begin{figure}[H]
\includegraphics[width=8cm,height=3cm]{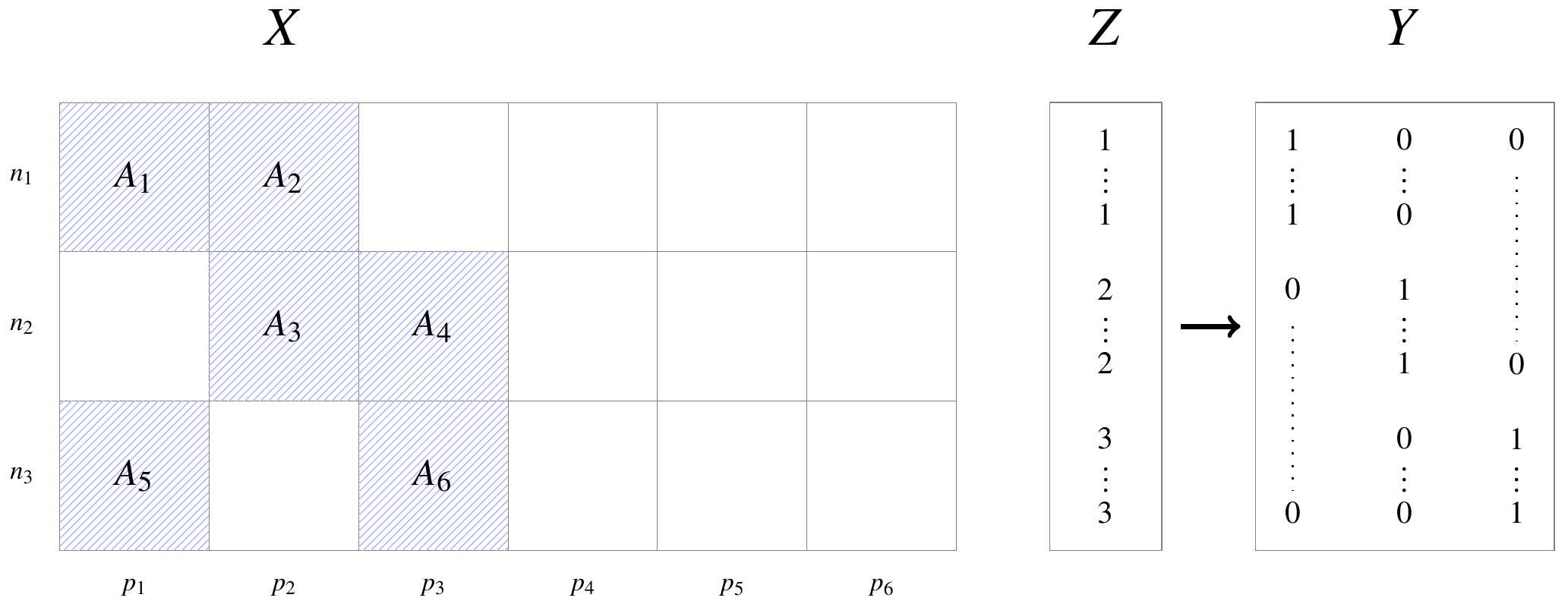}
\caption{Discriminant Analysis Design Matrices}\label{figdadesign}
\end{figure}

\begin{figure}[H]
\includegraphics[width=8cm,height=5cm]{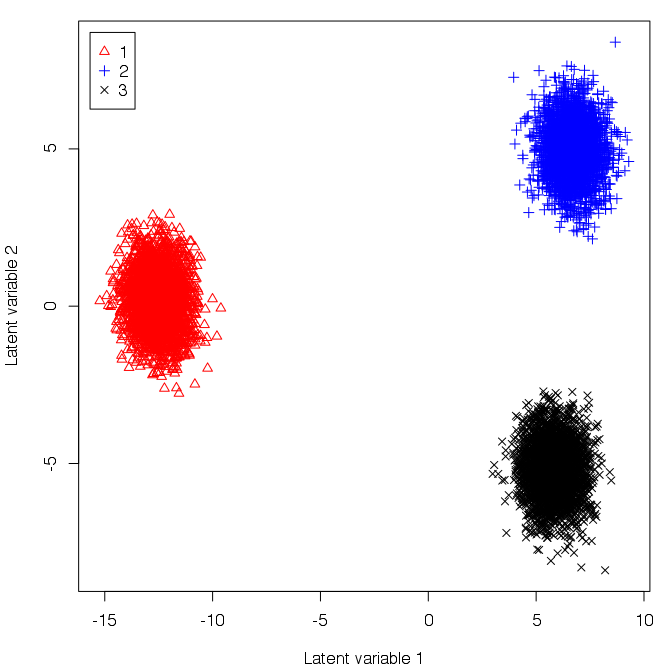}
\caption{Group PLS-DA on a big data set}\label{figda}
\end{figure}

We consider here the case of qualitative response variables for discrimination analysis. In this framework, PLS approaches have often been used \cite{Nguyen2002} by recoding the qualitative response as a dummy block matrix $\mat{Y}: n\times c$ indicating the class of each sample ($c$ being the number of categories). One can also directly apply PLS regression on the data as if $\mat{Y}$ was a matrix with continuous entries (from now on called PLS-DA). Note that \cite{Bar03} give some theoretical
justification for this approach. A group and a sparse group version have been proposed by \cite{Liquet2016} using only penalties on the loading related to the variables in $\mat{X}$.  
Our unified algorithm is then naturally extended in the same way to deal with categorical variables. We illustrate it on a big data set defined as follows.
Let $A_k$ be the set of indices $(i,j)$ of the $i$-th observation and $j$-th variable that are associated to the corresponding grey cell as shown in Figure~\ref{figdadesign}.
$\forall k=1,\dots,6,\ \forall i=1,\ldots,n , \forall j=1,\ldots,p$ $$X_{i,j}=\mu_k\times 1_{\{(i,j)\in A_k\}}+\epsilon_{i,j}$$ where $\vect{\mu}\transp=(\mu_1,\ldots,\mu_6)=(-1.0, 1.5, 1.0, 2.5, -0.5, 2.0)$, and $\epsilon_{i,j} \sim N(0,1)$. As illustrated on Figure~\ref{figdadesign}, the matrix $\mat{X}$ is composed of 6 groups of $p_k=100$ variables ($p=\sum_{k=1}^6p_k=600$) and each of the 3 categories of the response variable are linked to two groups of variables.
We used a sample size of $n=486,000$  which corresponds  to storage requirements of approximately 5~GB for the $\mat{X}$ matrix. We use $G=100$ chunks for computing the different matrix products. The run took around 9 minutes for a model using 2 components. 
The relevant groups have been selected in both components. We randomly sample $9,000$ observations and present in Figure~\ref{figda} their projection on the two components estimated on the full data set.  A nice discrimination of the 3 categories of the response variable is observed.

\section{Conclusion and Future Work}\label{Conclusion}

This paper surveys four popular partial least squares methods, and unifies these methods with recent variable selection techniques based on penalised singular value decomposition. We present a general framework for both symmetric and asymmetric penalised PLS methods and showcase some possible convex penalties. A unified algorithm is described and implemented for the penalised PLS methods, and we offer further extensions to deal with massive data sets ($n$, $p$ and $q$  very large).  A full comparison in terms of time and memory of the different proposed extensions is an open area of future research.

Aside from computational issues, it is unclear if retaining the deflations of the usual PLS methods is appropriate when there is penalisation. In particular, we note that the orthogonality constraints of the original PLS methods are not retained for the penalised methods. Further development of our methods could seek to preserve the orthogonality constraints. We are perusing this open area using ideas from \cite{tibshirani2011}, and \cite{Mackey08} for the simple lasso penalty. However, further investigation is required in the context of more complex penalties such as group or sparse group penalties.


%

\appendices
\section{Proofs of some results}\label{Proofs}

\subsection{Proof of (C1) in subsection~\ref{linkSVDtocovcor}\label{C1Proof}}

The proof is given here for completeness. It follows the lines of \cite[example Sec. 2.4]{Hsieh2009}.

Imposing $\|\vect{u}\|_2^{} = \|\vect{v}\|_2^{} = 1$ we have
\begin{eqnarray*}
\|\mat{M} - \delta\vect{u}\vect{v}\transp\|_F^2 = \|\mat{M}\|_F^2 - 2\delta\vect{u}\transp\mat{M}\vect{v} + \delta^2.
\end{eqnarray*}
Since $\mat{M}$ is fixed, the minimisation problem  is equivalent to
\begin{eqnarray*}
\underset{\|\vect{u}\|_2^{} = \|\vect{v}\|_2^{} = 1,~ \delta>0}{\text{minimise}}~- 2\delta\vect{u}\transp\mat{M}\vect{v} + \delta^2
\end{eqnarray*}
subject to $\vect{u}\transp\vect{u}_j = \vect{v}\transp\vect{v}_j = 0, 1\leq j<h$. 
The Lagrangian is
\begin{eqnarray*}
L=&-2\delta\vect{u}\transp\mat{M}\vect{v} +\delta^2 - \alpha(\vect{u}\transp\vect{u}-1)
-\beta(\vect{v}\transp\vect{v}-1)\\ & -\sum_{j=1}^{h-1}\mu_j\vect{u}\transp\vect{u}_j -\sum_{j=1}^{h-1}\nu_j\vect{v}\transp\vect{v}_j
\end{eqnarray*}
with Lagrangian multipliers $\alpha, \beta, \mu_j, \nu_j$ for $1 \leq j <h$. Now
\begin{eqnarray*}
\frac{\partial}{\partial\delta}L=&-2\vect{u}\transp\mat{M}\vect{v} +2\delta
\end{eqnarray*}
which should be equal to 0 at the optimum, leading to
\begin{eqnarray*}
\delta=&\vect{u}\transp\mat{M}\vect{v}.
\end{eqnarray*}
Substituting this $\delta$ into the optimisation function gives
\begin{eqnarray*}
\underset{\|\vect{u}\|_2^{} = \|\vect{v}\|_2^{} = 1}{\text{minimise}}~- (\vect{u}\transp\mat{M}\vect{v})^2
\end{eqnarray*}
subject to $\vect{u}\transp\vect{u}_j = \vect{v}\transp\vect{v}_j = 0, 1\leq j<h$. Noting that $\vect{u}\transp\mat{M}\vect{v} = Cov(\mat{X}\vect{u}, \mat{Y}\vect{v})$ and since we impose $\delta>0$, this can be rewritten as
\begin{eqnarray*}
\underset{\|\vect{u}\|_2^{} = \|\vect{v}\|_2^{} = 1}{\text{maximise}}~Cov(\mat{X}\vect{u}, \mat{Y}\vect{v})
\end{eqnarray*}
subject to $\vect{u}\transp\vect{u}_j = \vect{v}\transp\vect{v}_j = 0, 1\leq j<h$. We now consider the claim for the first pair of singular vectors 
\begin{eqnarray*}
&\underset{\|\vect{u}\|_2^{} = \|\vect{v}\|_2^{} = 1}{\text{maximise}}~Cov(\mat{X}\vect{u}, \mat{Y}\vect{v}),
\end{eqnarray*}
with Lagrangian
\begin{eqnarray*}
L=&\vect{u}\transp\mat{X}\transp\mat{Y}\vect{v} + \alpha(\vect{u}\transp\vect{u}-1)
+\beta(\vect{v}\transp\vect{v}-1).
\end{eqnarray*}
We have to solve
$$
\left\{\begin{array}{lclcl}
\frac{\partial}{\partial\vect{u}}L & = & \mat{X}\transp\mat{Y}\vect{v}+2\alpha\vect{u} & = & 0\\
\frac{\partial}{\partial\vect{v}}L & = & \mat{Y}\transp\mat{X}\vect{u}+2\beta\vect{v} & = & 0\\
\frac{\partial}{\partial\alpha}L & = & \vect{u}\transp\vect{u}-1 & = & 0\\
\frac{\partial}{\partial\beta}L & = & \vect{v}\transp\vect{v}-1 & = & 0
\end{array}\right.
$$
(Note that $\vect{v}$ is proportional to $\mat{Y}\transp\mat{X}\vect{u}$.) We multiply the first equation by $\vect{u}\transp$ and the second by $\vect{v}\transp$. This gives, using the third and the fourth,
$$
\alpha = \beta = -2^{-1}\vect{u}\transp\mat{X}\transp\mat{Y}\vect{v} = -2^{-1}Cov(\mat{X}\vect{u},\mat{Y}\vect{v}).
$$
We multiply the first equation by $\mat{Y}\transp\mat{X}$ and the second by $\mat{X}\transp\mat{Y}$. This gives
$$
\mat{Y}\transp\mat{X}\mat{X}\transp\mat{Y}\vect{v}+2\alpha\mat{Y}\transp\mat{X}\vect{u}=0
$$
and thus 
$$
(\mat{X}\transp\mat{Y})\transp\mat{X}\transp\mat{Y}\vect{v}=4\alpha\beta\vect{v}.
$$
Similarly
$$
(\mat{Y}\transp\mat{X})\transp\mat{Y}\transp\mat{X}\vect{u}=4\alpha\beta\vect{u}.
$$
So $\vect{u}_1$ and $\vect{v}_1$ are (normed) eigenvectors respectively of $(\mat{Y}\transp\mat{X})\transp\mat{Y}\transp\mat{X}$ and $(\mat{X}\transp\mat{Y})\transp\mat{X}\transp\mat{Y}$ associated to the same eigenvalue ($\lambda=4\alpha\beta$). Now,
$$
Cov^2(\mat{X}\vect{u},\mat{Y}\vect{v}) = \lambda,
$$
so $\vect{u}_1$ and $\vect{v}_1$ must be the eigenvectors associated to the largest eigenvalue, noted $\lambda_1$. Also, we have to choose the sign of $\vect{u}_1$ (or $\vect{v}_1$) so that the covariance is maximal and positive. Now, for the remaining $\vect{u}_h$ and $\vect{v}_h$ ($h>1$), since they must also maximize the covariance (under some successive added orthogonality constraints), they also need to be eigenvectors associated to the same matrices $(\mat{Y}\transp\mat{X})\transp\mat{Y}\transp\mat{X}$ and $(\mat{X}\transp\mat{Y})\transp\mat{X}\transp\mat{Y}$. It is clear that they are the eigenvectors associated to the remaining eigenvalues $\lambda_2>\cdots>\lambda_{\min(p,q)}$, and that
$$
\sqrt{\lambda_h} = Cov(\mat{X}\vect{u}_h,\mat{Y}\vect{v}_h)
$$
if the sign of $\vect{u}_h$ (or $\vect{v}_h$) is set correctly. It is easy to conclude using the link between the SVD and the eigen decomposition that $\vect{u}_h$ and $\vect{v}_h$ are the singular vectors of $\mat{X}\transp\mat{Y}$.

\subsection{Proof of (C2) in subsection~\ref{linkSVDtocovcor}\label{C2Proof}}

We want to find the successive pairs of vectors $(\tilde{\vect{w}}_1,\tilde{\vect{z}}_1),\ldots,(\tilde{\vect{w}}_r,\tilde{\vect{z}}_r)$ solution of
$$
\underset{\tilde{\vect{w}},\tilde{\vect{z}}}{\argmax}~Cor(\mat{X}\tilde{\vect{w}},\mat{Y}\tilde{\vect{z}}),
$$
subject to the constraints $Cov(\mat{X}\tilde{\vect{w}},\mat{X}\tilde{\vect{z}}_j)=Cov(\mat{Y}\tilde{\vect{w}},\mat{Y}\tilde{\vect{z}}_j)=0$, $1\leq j<h$. 

Let $\vect{u} = (\mat{X}\transp\mat{X})^{1/2}\tilde{\vect{w}}$ and $\vect{v} = (\mat{Y}\transp\mat{Y})^{1/2}\tilde{\vect{z}}$. We have
\begin{eqnarray*}
Cor(\mat{X}\tilde{\vect{w}},\mat{Y}\tilde{\vect{z}}) & = & \frac{\tilde{\vect{w}}\transp\mat{X}\transp\mat{Y}\tilde{\vect{z}}}{\sqrt{(\tilde{\vect{w}}\transp\mat{X}\transp\mat{X}\tilde{\vect{w}})(\tilde{\vect{z}}\transp\mat{Y}\transp\mat{Y}\tilde{\vect{z}})}}\\
& = & \frac{\vect{u}\transp(\mat{X}\transp\mat{X})^{-1/2}\mat{X}\transp\mat{Y}(\mat{Y}\transp\mat{Y})^{-1/2}\vect{v}}{\sqrt{(\vect{u}\transp\vect{u})(\vect{v}\transp\vect{v})}}.
\end{eqnarray*}
Since the above expression is invariant to the scaling of $\vect{u}$ and $\vect{v}$, the objective function is equivalent to maximising 
the \textit{covariance} between the scores under the constraint that their variances is equal to 1. This is also equivalent to maximising 
$$
\underset{\|\vect{u}\|_2=\|\vect{v}\|_2=1}{\argmax}~Cov(\mat{X}(\mat{X}\transp\mat{X})^{-1/2}\vect{u},\mat{Y}(\mat{Y}\transp\mat{Y})^{-1/2}\vect{v}),
$$
subject to the constraints
\begin{eqnarray*}
Cov(\mat{X}(\mat{X}\transp\mat{X})^{-1/2}\vect{u},\mat{X}(\mat{X}\transp\mat{X})^{-1/2}\vect{u}_j) & = & \\
Cov(\mat{Y}(\mat{Y}\transp\mat{Y})^{-1/2}\vect{v},\mat{Y}(\mat{Y}\transp\mat{Y})^{-1/2}\vect{v}_j) & = & 0, 
\end{eqnarray*}
$1\leq j<h$. But note that
\begin{eqnarray*}
Cov(\mat{X}(\mat{X}\transp\mat{X})^{-1/2}\vect{u}_h,\mat{X}(\mat{X}\transp\mat{X})^{-1/2}\vect{u}_j) &  &\\
= \vect{u}_h\transp(\mat{X}\transp\mat{X})^{-1/2}\mat{X}\transp\mat{X}(\mat{X}\transp\mat{X})^{-1/2}\vect{u}_j & = & \vect{u}_h\transp\vect{u}_j.
\end{eqnarray*}
and similarly for $\vect{v}$. So we in fact want to solve
$$
\underset{\|\vect{u}\|_2=\|\vect{v}\|_2=1}{\argmax}~Cov(\mat{X}(\mat{X}\transp\mat{X})^{-1/2}\vect{u},\mat{Y}(\mat{Y}\transp\mat{Y})^{-1/2}\vect{v}),
$$
subject to the constraints $\vect{u}\transp\vect{u}_j = \vect{v}\transp\vect{v}_j = 0$, $1\leq j<h$. 
Applying (C1), it is direct that they are the singular vectors of $(\mat{X}\transp\mat{X})^{-1/2}\mat{X}\transp\mat{Y}(\mat{Y}\transp\mat{Y})^{-1/2}$.\\

\subsection{Link between eigen elements and singular elements~\label{linkEigenElements}}

Let
$$
\mat{X}=\mat{U}\mat{D}\mat{V}\transp
$$
be the singular decomposition of some matrix $\mat{X}$. Now,
\begin{eqnarray*}
\mat{X}\transp\mat{X} & = & \mat{V}\mat{D}\mat{U}\transp\mat{U}\mat{D}\mat{V}\transp\\
& = & \mat{V}\mat{D}^2\mat{V}\transp.
\end{eqnarray*}
We recognize the eigenvalue decomposition of the matrix $\mat{X}$. Thus, it is clear that the 
eigenvalues of $\mat{X}\transp\mat{X}$ are the squares of the singular values of $\mat{X}$, and that the eigenvectors of $\mat{X}\transp\mat{X}$ are the right singular vectors of $\mat{X}$. Similarly for the left eigenelements:
\begin{eqnarray*}
\mat{X}\mat{X}\transp & = & \mat{U}\mat{D}\mat{V}\transp\mat{V}\mat{D}\mat{U}\transp\\
& = & \mat{U}\mat{D}^2\mat{U}\transp.
\end{eqnarray*}


\subsection{The two versions of NIPALS: scaled/unscaled~\label{TwoNIPALSversions}}

From the (compact) SVD decomposition $\mat{M}_{h-1}=\mat{X}_{h-1}\transp\mat{Y}_{h-1}=\mat{U}_h\mat{\Delta}_h\mat{V}_h\transp$, we obtain $\mat{V}_h=\mat{Y}_{h-1}\transp\mat{X}_{h-1}\mat{U}_{h}\mat{\Delta}_{h}^{-1}$ and thus $\vect{v}_h =\delta_h^{-1}\mat{Y}_{h-1}\transp\vect{\xi}_h = (\|\vect{\xi}_h\|^{-2}\delta_h)^{-1}\mat{Y}_{h-1}\transp\vect{\xi}_h/\|\vect{\xi}_h\|^2$, where $\vect{v}_h$ is the first column of $\mat{V}_h$ and $\delta_h=\|\mat{Y}_{h-1}\transp\vect{\xi}_h\|$ is the first diagonal element of $\mat{\Delta}_h$. This vector $\vect{v}_h$ is normed. This is exactly was is done in \cite[p. 212, step 6]{Hoskuldsson1988} (despite an erroneous transpose sign). But this differs to the classic PLS2 algorithm \cite{Wold1984} which follows the same process but does not include this scaling; see \cite[p. 117]{Wold2001} or \cite[p. 128]{Tenenhaus1998}. They instead compute, at each step $h$, a (not scaled) vector $\mat{Y}_{h-1}\transp\vect{\xi}_h/(\vect{\xi}_h\transp\vect{\xi}_h)$, which they note $\vect{c}$ (not to be confounded with our $c_h$). It is proportional to our $\vect{v}_h$, with $\vect{v}_h = (\vect{c}\transp\vect{c})^{-1/2}\vect{c} = \alpha_h\vect{c}$, where $\alpha_h := \|\vect{\xi}_h\|^2 / \|\mat{Y}_{h-1}\transp\vect{\xi}_h\|$. \\

Now, define $p_h=\alpha_h^{-1}$ in the scaled case and $p_h=1$ otherwise. The $Y$-score vectors are defined as $\vect{\omega}_h = p_h\alpha_h\mat{Y}_{h-1}\vect{v}_h$ (which is noted $\vect{u}$ by the authors of the unscaled case).\\

For both algorithms, the fitted values $\widehat{\mat{Y}}_{(h)} = b_h\vect{\xi}_h\vect{v}_h\transp$ (or $\widehat{\mat{Y}}_{(h)} = b_h\vect{\xi}_h\vect{c}\transp$ for the unscaled case) are computed at each step $h$, where $b_h = \vect{\omega}_h\transp\vect{\xi}_h/(\vect{\xi}_h\transp\vect{\xi}_h)$ is the coefficient when you regress $\vect{\omega}_h$ on $\vect{\xi}_h$ (and is at the core of the inner relation explicited in the next subsection). One can show that $b_h=p_h$ and that $\widehat{\mat{Y}}_{(h)} = \mathcal{P}_{\vect{\xi}_h}\mat{Y}_{h-1}$. Indeed, since $\vect{v}_h = \mat{Y}_{h-1}\transp\vect{\xi}_h/\|\mat{Y}_{h-1}\transp\vect{\xi}_h\|$, we obtain
\begin{eqnarray*}
\vect{\omega}_h\transp \vect{\xi}_h & = & p_h\alpha_h\vect{v}_h\transp\mat{Y}_{h-1}\transp\vect{\xi}_h \\
& = & p_h\alpha_h\|\mat{Y}_{h-1}\transp\vect{\xi}_h\|\\
& = & p_h\|\vect{\xi}_h\|^2.
\end{eqnarray*}

For scaled weights, we have
\begin{eqnarray*}
\widehat{\mat{Y}}_{(h)} & = & b_h\vect{\xi}_h\vect{v}_h\transp\\
& = & \frac{\|\mat{Y}_{h-1}\transp\vect{\xi}_h\|}{(\vect{\xi}_h\transp\vect{\xi}_h)}\vect{\xi}_h\frac{\vect{\xi}_h\transp\mat{Y}_{h-1}}{\|\mat{Y}_{h-1}\transp\vect{\xi}_h\|}\\
& = & \vect{\xi}_h(\vect{\xi}_h\transp\vect{\xi}_h)^{-1}\vect{\xi}_h\transp\mat{Y}_{h-1}\\
& = & \mathcal{P}_{\vect{\xi}_h}\mat{Y}_{h-1}.
\end{eqnarray*}
For unscaled weights, we have also
$$
\widehat{\mat{Y}}_{(h)} = b_h\vect{\xi}_h\vect{c}\transp = 1\cdot\vect{\xi}_h(\vect{\xi}_h\transp\vect{\xi}_h)^{-1}\vect{\xi}_h\transp\mat{Y}_{h-1} = \mathcal{P}_{\vect{\xi}_h}\mat{Y}_{h-1}.
$$

As in case (ii), using \cite[Theorem~7, p.~151]{Puntanen2011} ,
we obtain
\begin{align*}
\widehat{\mat{Y}}_{(h)} &= \mathcal{P}_{\vect{\xi}_h}\mat{Y}_{h-1}
=\mathcal{P}_{\vect{\xi}_h}(\mat{I} - \mathcal{P}_{\vect{\Xi}_{\bullet h-1}})\mat{Y}
=\mathcal{P}_{\vect{\xi}_h}\mat{Y}
\end{align*}
and
$$
\widehat{\mat{Y}}_{h} = \sum_{j=1}^h\widehat{\mat{Y}}_{(j)} = \mathcal{P}_{\vect{\Xi}_{\bullet h}}\mat{Y}.
$$

\subsection{Proof of the inner relation in PLS-R~\label{ProofInner}}

The central inner PLS relation is made of successive \textit{univariate} regressions of $\vect{\omega}_h$ upon $\vect{\xi}_h$. This constitute the link between $Y$ and $X$ in the PLS model. This link is estimated one dimension at a time (partial modeling) hence the original ``Partial'' in the PLS acronym.

We have
\begin{eqnarray*}
\vect{\omega}_h & = & \mathcal{P}_{\vect{\xi}_h}\vect{\omega}_h + \mathcal{P}_{\vect{\xi}_h^\perp}\vect{\omega}_h\\
& = &  \vect{\xi}_h(\vect{\xi}_h\transp\vect{\xi}_h)^{-1}\vect{\xi}_h\transp\vect{\omega}_h + \mathcal{P}_{\vect{\xi}_h^\perp}\vect{\omega}_h\\
& = & \vect{\xi}_hp_h + \mathcal{P}_{\vect{\xi}_h^\perp}\vect{\omega}_h\\
& := & \vect{\xi}_hp_h + \vect{r}_h.
\end{eqnarray*}
This leads to
\begin{eqnarray*}
\mat{\Omega}_{\bullet h} & = & \mat{\Xi}_{\bullet h}\mat{P}_{h} + \mat{R}_{\bullet h},
\end{eqnarray*}
where $\mat{P}_h=\textrm{diag}(p_j)_{1\leq j\leq h}$ and $\mat{R}_H = [\vect{r}_1,\ldots,\vect{r}_H]$.

\subsection{The decomposition model in PLS-R~\label{DecompositionPLSR}}

Note that due to the properties on the $\vect{\xi}_j$, we have that $\mat{\Xi}_{H}\transp\mat{\Xi}_{H}$ is a diagonal matrix and also that $\vect{\xi}_h\transp=\vect{\xi}_h\transp\left(\prod_{j=h-1}^{1}\mathcal{P}_{\vect{\xi}_j^\perp}\right)=\vect{\xi}_h\transp\mathcal{P}_{\mat{\Xi}_{\bullet h-1}^\perp}$. This allows us to write
\begin{eqnarray*}
\mat{C}_{H}\transp & := & (\mat{\Xi}_{H}\transp\mat{\Xi}_{H})^{-1}\mat{\Xi}_{H}\transp\mat{X} \\
& = & (\mat{\Xi}_{H}\transp\mat{\Xi}_{H})^{-1}\left[
\begin{array}{c}
\vect{\xi}_1\transp\\
\vect{\xi}_2\transp\mathcal{P}_{\mat{\Xi}_{\bullet 1}^\perp}\\
\vdots\\
\vect{\xi}_H\transp\mathcal{P}_{\mat{\Xi}_{\bullet H-1}^\perp}\\
\end{array}
\right]\mat{X}\\
& = & (\mat{\Xi}_{H}\transp\mat{\Xi}_{H})^{-1}\left[
\begin{array}{c}
\vect{\xi}_1\transp\mat{X}_0\\
\vect{\xi}_2\transp\mat{X}_1\\
\vdots\\
\vect{\xi}_H\transp\mat{X}_{H-1}\\
\end{array}
\right]\\
& = & \left[
\begin{array}{c}
(\vect{\xi}_1\transp\vect{\xi}_1)^{-1}\vect{\xi}_1\transp\mat{X}_0\\
(\vect{\xi}_2\transp\vect{\xi}_2)^{-1}\vect{\xi}_2\transp\mat{X}_1\\
\vdots\\
(\vect{\xi}_H\transp\vect{\xi}_H)^{-1}\vect{\xi}_H\transp\mat{X}_{H-1}\\
\end{array}
\right],
\end{eqnarray*}

Since $\vect{v}_h$ is normed, then $\mathcal{P}_{\vect{v}_h} = \vect{v}_h\vect{v}_h\transp$. Now, looking more closely at the PLS-R algorithm (see, e.g., \cite[p.~3]{Vinzi2013}, \cite{Abdi2007}), it is clear that
$$
\vect{\omega}_h = \mat{Y}_{h-1}\vect{v}_h
$$
and
$$
\mat{Y}_{h-1} = \mathcal{P}_{\vect{\xi}_h^\perp}\mat{Y}_{h-1} + \mathcal{P}_{\vect{\xi}_h}\mat{Y}_{h-1} = \mat{Y}_h + \mathcal{P}_{\vect{\xi}_h}\mat{Y}_{h-1}.
$$
We thus have
\begin{eqnarray*}
\mat{Y}_{h-1} & = & \mat{Y}_{h-1} \mathcal{P}_{\vect{v}_{h}} + \mat{Y}_{h-1} \mathcal{P}_{\vect{v}_{h}^\perp} \\
& = & \vect{\omega}_h\vect{v}_h\transp + \mat{Y}_{h-1} - \mat{Y}_{h-1}\mathcal{P}_{\vect{v}_{h}} \\
& = & \vect{\omega}_h\vect{v}_h\transp + \mat{Y}_h + \mathcal{P}_{\vect{\xi}_h}\mat{Y}_{h-1} - \mat{Y}_{h-1}\mathcal{P}_{\vect{v}_{h}}.
\end{eqnarray*}
By recurrence, we obtain
\begin{eqnarray*}
\mat{Y}_0 & = & \sum_{j=1}^{H}\left[\vect{\omega}_j\vect{v}_j\transp + \mathcal{P}_{\vect{\xi}_j}\mat{Y}_{j-1} - \mat{Y}_{j-1}\mathcal{P}_{\vect{v}_j}\right] + \mat{Y}_H\\
& = & \mat{\Omega}_H\mat{D}_H\transp + \sum_{j=1}^{H}\left[\vect{\xi}_j(\vect{\xi}_j\transp\vect{\xi}_j)^{-1}\vect{\xi}_j\transp\mat{Y}_{j-1} - \mat{Y}_{j-1}\vect{v}_j\vect{v}_j\transp\right] + \mat{Y}_H\\
& = & \mat{\Omega}_H\mat{D}_H\transp + \sum_{j=1}^{H}\left[\vect{\xi}_j\vect{g}_j\transp - \vect{\omega}_j\vect{v}_j\transp\right] + \mat{Y}_H\\
& = & \mat{\Omega}_H\mat{D}_H\transp + \mat{\Xi}_H\mat{G}_H\transp - \mat{\Omega}_H\mat{D}_H\transp + \mat{Y}_H,
\end{eqnarray*}
where we have defined $\mat{D}_H=[\vect{v}_1,\ldots,\vect{v}_H]$ and where
\begin{eqnarray*}
\mat{G}_{H}\transp & := & \left[
\begin{array}{c}
(\vect{\xi}_1\transp\vect{\xi}_1)^{-1}\vect{\xi}_1\transp\mat{Y}_0\\
(\vect{\xi}_2\transp\vect{\xi}_2)^{-1}\vect{\xi}_2\transp\mat{Y}_1\\
\vdots\\
(\vect{\xi}_H\transp\vect{\xi}_H)^{-1}\vect{\xi}_H\transp\mat{Y}_{H-1}.
\end{array}
\right].
\end{eqnarray*}

\begin{remark}
Let 
\begin{eqnarray*}
\widetilde{\mat{G}}_{H}\transp & := & \left[
\begin{array}{c}
(\|\mat{Y}_0\transp\vect{\xi}_1\|)^{-1}\vect{\xi}_1\transp\mat{Y}_0\\
(\|\mat{Y}_1\transp\vect{\xi}_2\|)^{-1}\vect{\xi}_2\transp\mat{Y}_1\\
\vdots\\
(\|\mat{Y}_{H-1}\transp\vect{\xi}_H\|)^{-1}\vect{\xi}_H\transp\mat{Y}_{H-1}\\
\end{array}
\right]\\
& = & \mat{P}_H^{-1}\mat{G}_{H}\transp,
\end{eqnarray*}
where $\mat{P}_H=(p_h)$ is the diagonal matrix defined in subsection~\ref{ProofInner}. It can be seen to have entries $\|\mat{Y}_{h-1}\transp\vect{\xi}_h\| / (\vect{\xi}_h\transp\vect{\xi})$, $h=1,\ldots,H$. We have
\begin{eqnarray*}
\mat{Y}_h & = & \mat{Y}_{h-1}-\vect{\xi}_h\vect{g}_h\transp\\
& = & \mat{Y}_{h-1}-p_h\vect{\xi}_h\tilde{\vect{g}}_h\transp,
\end{eqnarray*}
the first deflation step formula being the one used in \cite{Tenenhaus1998} while the second is the one used in \cite{Hoskuldsson1988}; see also \cite{Vinzi2013}. 
\end{remark}

We have \cite[p. 101, g)]{Tenenhaus1998}
\begin{eqnarray*}
\mat{X}_h & = & \mathcal{P}_{\vect{\xi}_h^\perp}\mat{X}_{h-1}\\
& = & \left(\mat{I} - \vect{\xi}_h(\vect{\xi}_h\transp\vect{\xi}_h)^{-1}\vect{\xi}_h\transp\right)\mat{X}_{h-1}\\
& = & \left(\mat{I} - \mat{X}_{h-1}\vect{u}_h(\vect{\xi}_h\transp\vect{\xi}_h)^{-1}\vect{\xi}_h\transp\right)\mat{X}_{h-1}\\
& = & \mat{X}_{h-1}\left(\mat{I} - \vect{u}_h(\vect{\xi}_h\transp\vect{\xi}_h)^{-1}\vect{\xi}_h\transp\mat{X}_{h-1}\right)\\
& = & \mat{X}_{h-2}\left(\mat{I} - \vect{u}_{h-1}(\vect{\xi}_{h-1}\transp\vect{\xi}_{h-1})^{-1}\vect{\xi}_{h-1}\transp\mat{X}_{h-2}\right)\\
& &\times\left(\mat{I} - \vect{u}_h(\vect{\xi}_h\transp\vect{\xi}_h)^{-1}\vect{\xi}_h\transp\mat{X}_{h-1}\right)\\
& = & \mat{X}_0\prod_{j=1}^{h}\left(\mat{I} - \vect{u}_j(\vect{\xi}_j\transp\vect{\xi}_j)^{-1}\vect{\xi}_j\transp\mat{X}_{j-1}\right)\\
& = & \mat{X}_0\prod_{j=1}^{h}\left(\mat{I} - \vect{u}_j(\vect{\xi}_j\transp\vect{\xi}_j)^{-1}\vect{\xi}_j\transp\mathcal{P}_{\mat{\Xi}_{\bullet j-1}^\perp}\mat{X}_0\right)\\
& = & \mat{X}_0\prod_{j=1}^{h}\left(\mat{I} - \vect{u}_j(\vect{\xi}_j\transp\vect{\xi}_j)^{-1}\vect{\xi}_j\transp(\mat{I} - \mathcal{P}_{\mat{\Xi}_{\bullet j-1}})\mat{X}_0\right)\\
& = & \mat{X}_0\prod_{j=1}^{h}\left(\mat{I} - \vect{u}_j(\vect{\xi}_j\transp\vect{\xi}_j)^{-1}\vect{\xi}_j\transp\mat{X}_0\right) \\
& := & \mat{X}_0\mat{A}^{(h)}
\end{eqnarray*}
so
\begin{eqnarray*}
\mat{\Xi}_{\bullet h} & = & [\vect{\xi}_1,\ldots,\vect{\xi}_h]\\
& = & [\mat{X}\vect{u}_1,\ldots,\mat{X}_{h-1}\vect{u}_h]\\
& = & [\mat{X}\vect{u}_1,\ldots,\mat{X}\mat{A}^{(h-1)}\vect{u}_h]\\
& = & \mat{X}[\vect{u}_1,\ldots,\mat{A}^{(h-1)}\vect{u}_h]\\
\end{eqnarray*}

Let the matrix of adjusted weights be $\tilde{\mat{W}}_{\bullet h} = [\tilde{\vect{w}}_1,\ldots,\tilde{\vect{w}}_h]$ with
$\tilde{\vect{w}}_1 = \vect{u}_1$ and $\tilde{\vect{w}}_h = \prod_{j=1}^{h-1}(\mat{I} - \vect{u}_j\vect{c}_j\transp)\vect{u}_h=\prod_{j=1}^{h-1}(\mat{I} - \vect{u}_j(\vect{\xi}_j\transp\vect{\xi}_j)^{-1}\vect{\xi}_j\transp\mat{X})\vect{u}_h=\mat{A}^{(h-1)}\vect{u}_h$. It is thus clear that $\mat{X}\vect{w}_h = \mat{X}_{h-1}\vect{u}_h$ and $\mat{\Xi}_{\bullet h} = \mat{X}\tilde{\mat{W}}_{\bullet h}$; see \cite[p. 135]{Tenenhaus1998}.
Interestingly, from \cite[p. 114]{Tenenhaus1998}, we can also write $\tilde{\mat{W}}_{\bullet h} = \mat{U}_{\bullet h}(\mat{C}_{\bullet h}\transp\mat{U}_{\bullet h})^{-1}$.

From \cite[Theorem~4, p.~106]{Puntanen2011},
\begin{align*}
\mat{C}_{H}\transp\tilde{\mat{W}}_{H}\mat{C}_{H}\transp &= (\mat{\Xi}_{H}\transp \mat{\Xi}_{H})^{-1}\mat{\Xi}_{H}\transp\mat{X}\tilde{\mat{W}}_{H}\mat{C}\transp_{H}\\
&= (\mat{\Xi}_{H}\transp \mat{\Xi}_{H})^{-1}\mat{\Xi}_{H}\transp\mat{\Xi}_{H}\mat{C}\transp_{H}\\
&= \mat{C}\transp_{H}
\end{align*}
so that $\tilde{\mat{W}}_{H}$ is a generalised inverse of $\mat{C}_{H}\transp$ \cite{deJong1993}. 

Suppose that $\textrm{rank}(\mat{X}_0)=r\leq p$. We have $\vect{\xi}_1=\mat{X}_0\vect{u}_1\in \mathcal{I}(\mat{X})$ of dimension $r$ (as a combination of the columns of $\mat{X}_0$). Then we define $\mat{X}_1=\mathcal{P}_{\vect{\xi}_1^\perp}\mat{X}_0=\mathcal{P}_{\vect{\xi}_1^\perp\cap \mathcal{I}(\mat{X})}\mat{X}_0+\mathcal{P}_{\vect{\xi}_1^\perp\cap \mathcal{I}(\mat{X})^\perp}\mat{X}_0=\mathcal{P}_{\vect{\xi}_1^\perp\cap \mathcal{I}(\mat{X})}\mat{X}_0$. So the columns of $\mat{X}_1$ belong to $\mathcal{I}(\mat{X})\cap\{\vect{\xi}_1\}^{\perp}$, which is of dimension $r-1$. We iterate this process \cite[Sec. 5]{Abdi2012} until we obtain $\mat{X}_{r}$ which will be of rank $0$ (and so $\mat{X}_r=\mat{0}$). We thus have the (exact) decomposition when $H=r$:
$$
\mat{X} = \mat{\Xi}_{\bullet r}\mat{C}_{\bullet r}\transp.
$$
From \cite[eq. 2.22 p. 16]{Seber2008}, the columns of $\mat{\Xi}_{\bullet r}$ are linearly independent. 
From \cite[eq. 7.54(d) p. 139]{Seber2008}, $\mat{\Xi}_{\bullet r}^+=(\mat{\Xi}_{\bullet r}\transp\mat{\Xi}_{\bullet r})^{-1}\mat{\Xi}_{\bullet r}\transp$ and $\mat{\Xi}_{\bullet r}^+\mat{\Xi}_{\bullet r}=\mat{I}_r$. So, we obtain
$$
\mat{\Xi}_{\bullet r}^+\mat{X}=\mat{C}_{\bullet r}\transp.
$$

\subsection{The adjusted weight optimisation problem~\label{AdjustedWeightProblem}}

Until now we have defined the X-scores $\vect{\xi}_h$ in terms of the deflated matrix $\mat{X}_h$, however, we can also define the scores using the original matrix $\mat{X}$ by a set of adjusted weight vectors $\tilde{\vect{w}}_h$ \cite{Braak1998}, as proved in the previous subsection:
\begin{align}
\label{eq:Wdef}
\vect{\xi}_h = \mat{X}\tilde{\vect{w}}_h = \mat{X}_{h-1}\vect{u}_h, \qquad h = 1,\dots H.
\end{align}
Let $\tilde{\mat{W}}_{\bullet h}$ denote the matrix with column vectors $\tilde{\vect{w}}_1,\dots, \tilde{\vect{w}}_h$ so that,
\begin{align*}
\mat{X} &= \mat{\Xi}_{\bullet h}\mat{C}_{\bullet h}\transp + \mat{X}_{h}\\
&= \mat{X}\tilde{\mat{W}}_{\bullet h}\mat{C}_{\bullet h}\transp + \mat{X}_{h}.
\end{align*}
Using the definition (\ref{eq:Wdef}) for any $h > 1$, and rearranging the above decomposition, we can write:
\begin{align*}
\mat{\xi}_h &= \mat{X}_{h-1}\vect{u}_h = \mat{X}(\mat{I}_p - \tilde{\mat{W}}_{\bullet h-1}\mat{C}_{\bullet h-1}\transp)\vect{u}_h
\end{align*}
and thus we can define the adjusted weights as:
\begin{align}
\label{eq:Clmsp}
\tilde{\vect{w}}_h &= (\mat{I}_p - \tilde{\mat{W}}_{\bullet h-1}\mat{C}_{\bullet h-1}\transp)\vect{u}_h.
\end{align}
Thus the adjusted weights can be found using the loadings and weights from previous iterations. Rearranging for $\vect{u}_h$ we have,
\begin{align}
\label{eq:weightu}
\begin{split}
\vect{u}_h &= \tilde{\vect{w}}_h + \tilde{\mat{W}}_{\bullet h-1}\mat{C}_{\bullet h-1}\transp\vect{u}_h\\
&= \tilde{\vect{w}}_h - \tilde{\mat{W}}_{\bullet h-1}\mat{g}_h
\end{split}
\end{align}
where $\mat{g}_h = -\mat{C}_{\bullet h-1}\transp\vect{u}_h$.

We have seen that $\tilde{\mat{W}}_{\bullet h}=\mat{U}_{\bullet h}(\mat{C}_{\bullet h}\transp\mat{U}_{\bullet h})^{-1}$, so that
$\tilde{\mat{W}}_{\bullet h}\transp\tilde{\mat{W}}_{\bullet h}=(\mat{C}_{\bullet h}\transp\mat{U}_{\bullet h}\mat{U}_{\bullet h}\transp\mat{C}_{\bullet h})^{-1}$ and $(\tilde{\mat{W}}_{\bullet h}\transp\tilde{\mat{W}}_{\bullet h})^{-1}$ exists. Consequently, $\tilde{\mat{W}}_{\bullet h}^+=(\tilde{\mat{W}}_{\bullet h}\transp\tilde{\mat{W}}_{\bullet h})^{-1}\tilde{\mat{W}}_{\bullet h}\transp=\mat{C}_{\bullet h}\transp\mat{U}_{\bullet h}\mat{U}_{\bullet h}\transp$ and $\tilde{\mat{W}}_{\bullet h}^+\tilde{\mat{W}}_{\bullet h}=\mat{I}$.

To express $\tilde{\vect{w}}_h$ in terms of $\vect{u}_h$ we first note that,
\begin{align*}
\tilde{\mat{W}}_{\bullet h-1}\vect{g}_h = \tilde{\vect{w}}_h - \vect{u}_h, 
\end{align*}
so that
\begin{align}
\label{eq:G}
\begin{split}
\mat{g}_h &= \tilde{\mat{W}}_{\bullet h-1}^+(\vect{w}_h - \vect{u}_h)\\
&= \tilde{\mat{W}}_{\bullet h-1}^+\vect{w}_h
\end{split}
\end{align}
where we use the fact that $\tilde{\mat{W}}_{\bullet h-1}^+\vect{u}_h = \vect{0}_{h-1}$ (since $\mat{U}_{\bullet h-1}\transp\vect{u}_h = \vect{0})$. 

Combining equations (\ref{eq:G}) and (\ref{eq:weightu}) gives:
\begin{align*}
\vect{u}_h &= (\mat{I}_p - \tilde{\mat{W}}_{\bullet h-1}\tilde{\mat{W}}_{\bullet h-1}^+)\tilde{\vect{w}}_h\\
&= \mathcal{P}_{\tilde{\mat{W}}_{\bullet h-1}^{\perp}}\tilde{\vect{w}}_h.
\end{align*}

\subsection{The sparse PLS weights~\label{sPLSWeights}}

The optimisation function for the $\tilde{\vect{u}}$ in sparse PLS is:
\begin{equation}\label{sPLS}
\begin{split}
\tilde{\vect{u}}_h = &\underset{\tilde{\vect{u}}}{\argmin}~\left\{\|\mat{M}_{h-1} - \tilde{\vect{u}}\vect{v}\transp\|_F^2 + 2\lambda_1\|\tilde{\vect{u}}\|_1\right\}.
\end{split}
\end{equation}
We denote $m_{ij,h}$ the entry $(i,j)$ of $\mat{M}_h$, $h=1,\ldots,H$. 
Solving this problem, we rewrite the criterion (\ref{sPLS}) as a separable function
$$\sum_{i=1}^p\left\{\sum_{j=1}^q(m_{ij}-\tilde{u}_{i}v_{j})^2+2\lambda_1|\tilde{u}_{i}|\right\}.$$
Therefore, we can optimise over individual components of $\tilde{\vect{u}}$ separately. Expanding the squares and observing that $||\vect{v}||_2=1$, we obtain
\begin{eqnarray*} 
\sum_{j=1}^q(m_{ij}-\tilde{u}_iv_j)^2&=&\sum_{j=1}^qm_{ij}^2-2\sum_{j=1}^qm_{ij}\tilde{u}_iv_j+\sum_{j=1}^q\tilde{u}_i^2v_j^2\\
&=&\sum_{j=1}^qm_{ij}^2-2(\mat{M}\vect{v})_i\tilde{u}_i+\tilde{u}_i^2,
\end{eqnarray*}
where $\mat{M}_h = (m_{ij})$. Hence, the optimal $\tilde{u}_i$ minimises $\tilde{u}_i^2-2(\mat{M}\vect{v})_i\tilde{u}_i+2\lambda_1|\tilde{u}_i|$. By using \cite[Lemma~2]{Shen2008}, we find
$$\tilde{u}_i = g^{\textrm{soft}}((\mat{M}\vect{v})_i,\lambda_1).$$
Similarly, optimisation  over $\tilde{\vect{v}}$ for a fixed (normed) $\vect{u}$ is also obtained by optimising over individual components:
$$\tilde{v}_j = g^{\textrm{soft}}((\mat{M}\transp\vect{u})_j,\lambda_2).$$
The minimiser of \eqref{sPLS} is obtained by applying the thresholding function $g^{\textrm{soft}}(\cdot,{\lambda})$ to the vector $\mat{M}\vect{v}$ componentwise and to the vector $\mat{M}\transp\vect{u}$ componentwise too.




\ifCLASSOPTIONcaptionsoff
  \newpage
\fi




\bibliographystyle{IEEEtran} 

\bibliography{biblio}
\end{document}